\documentclass[12pt, a4paper]{article}
\usepackage{amsfonts, amssymb}
\usepackage{graphicx}
\usepackage{latexsym,amsmath,color,array}

\usepackage{pifont} 

\usepackage{natbib}
\usepackage{hyperref}
\usepackage{enumitem}
\usepackage{bbding}
\usepackage{amssymb}
\usepackage{amsmath}
\usepackage{graphicx}
\usepackage{amsmath}
\usepackage{bbm}
\usepackage{mathtools}
 \usepackage{color}
 \usepackage{array}
 \usepackage{subcaption}
\usepackage{caption}
 \usepackage{multirow}
 \usepackage[dvipsnames]{xcolor}
 \usepackage{makecell}
 \usepackage{colortbl}
 \usepackage{algorithm}
 \usepackage{lmodern}
 \usepackage{tikz}
\usetikzlibrary{positioning}

\usepackage{float}
\usepackage{tikz}
\usepackage{booktabs}
\usepackage{pifont}
\usepackage{threeparttable}
\usepackage[T1]{fontenc}
\usepackage{lmodern}
\usetikzlibrary{decorations.pathreplacing,calligraphy}
\usepackage{standalone}
\usetikzlibrary{arrows.meta,positioning} 
\tikzset{>=Stealth}     
\usepackage{url}
\usepackage[T1]{fontenc}

\hypersetup{
    hidelinks = true
}
\usepackage{url}

  \usepackage{booktabs}
 \usepackage{threeparttable}
 \usepackage{longtable}
  \usepackage{subcaption}
\DeclareCaptionLabelFormat{andfigure}{#1~#2  \&  \figurename~\thefigure}

\def\1{\mathbb{I}}

\newcounter{appen}[section]

\begin{document}

\title{Comparison of generalised additive models and neural networks in applications: A systematic review}

\author{
\begin{tabular}{ccc}
\begin{tabular}[t]{@{}c@{}}Jessica Doohan\footnote{Department of Mathematics and Statistics, University of Limerick; doohan.jessica@ul.ie}\end{tabular} &
\begin{tabular}[t]{@{}c@{}}Lucas Kook\footnote{Institute for Statistics and Mathematics, Vienna University of Economics and Business; \\ 
\phantom{000.} lucasheinrich.kook@gmail.com}\end{tabular} &
\begin{tabular}[t]{@{}c@{}}Kevin Burke\footnote{Department of Mathematics and Statistics, University of Limerick;  kevin.burke@ul.ie}\end{tabular}
\end{tabular}
}

\date{\today}

\maketitle

\begin{abstract}
Neural networks have become a popular tool in predictive modelling, more commonly associated with machine learning and artificial intelligence than with statistics.
Generalised Additive Models (GAMs) are flexible non-linear statistical models that retain interpretability.
Both are state-of-the-art in their own right, with their respective advantages and disadvantages.
This paper analyses how these two model classes have performed on real-world tabular data.
Following PRISMA guidelines, we conducted a systematic review of papers that performed empirical comparisons of GAMs and neural networks. 
Eligible papers were identified, yielding 143 papers, with 430 datasets. 
Key attributes at both paper and dataset levels were extracted and reported.
Beyond summarising comparisons, we analyse reported performance metrics using mixed-effects modelling to investigate potential characteristics that can explain and quantify observed differences, including application area, study year, sample size, number of predictors, and neural network complexity. 
Across datasets, no consistent evidence of superiority was found for either GAMs or neural networks when considering the most frequently reported metrics (RMSE, $R^2$, and AUC). 
Neural networks tended to outperform in larger datasets and in those with more predictors, but this advantage narrowed over time. 
Conversely, GAMs remained competitive, particularly in smaller data settings, while retaining interpretability.
Reporting of dataset characteristics and neural network complexity was incomplete in much of the literature, limiting transparency and reproducibility.
This review highlights that GAMs and neural networks should be viewed as complementary approaches rather than competitors.
For many tabular applications, the performance trade-off is modest, and interpretability may favour GAMs.

\smallskip


\end{abstract}

\qquad

\newpage
\section{Introduction}
Over the past two decades, rapid advancements in machine learning and artificial intelligence have reshaped the abilities of predictive modelling.
Among the most transformative of these developments are neural networks \citep{RosenblattF1958TpAp, rumelhart1986learning}, which have demonstrated high predictive performance across a wide range of tasks, including image recognition, natural language processing, and forecasting \citep{abiodun2018state, liu2017survey}.
More recently, deep and alternative neural network architectures have been successful in a wide variety of tasks \citep{arik2021tabnet, hollmann2022tabpfn,zhou2024deep, jia2024performance, hollmann2025accurate, CNN_tab_AnH_ILDO}.
Notwithstanding these advances, Multilayer Perceptrons (MLPs) remain the most common neural network baseline for tabular datasets \citep{gorishniy2021revisiting}, and this is also confirmed within the large scale systematic review of the current paper.

In many cases, the primary objective when employing neural networks is to maximise predictive accuracy, with less emphasis placed on model interpretability. 
These models are often regarded as ``black-box'' algorithms: although their interconnected and non-linear structure allows for high flexibility, it also makes for a model that is difficult to interpret. 
One of the key strengths of neural networks is that, even with a single hidden layer, they have been proven to be universal approximators \citep{cybenko1989approximation, hornik1989multilayer}. 
This property implies that neural networks can approximate any continuous, single-valued function, and hence could be strong predictors.
However, these theories of universality may require the neural network size to grow to infinity.
Subsequent studies have shown that deep neural networks of a fixed size, i.e., neural networks with more than one hidden layer, can also achieve universal approximation, often with a drastically reduced number of parameters \citep{lin2017does, mhaskar2016deep}, thereby improving computational efficiency. Moreover, \cite{choromanska2015loss} show that the loss landscape of deeper networks contains many local minima, which are not difficult to find using stochastic gradient descent, and perform similarly to the global minimum. Interestingly, for regression tasks, \cite{yun2019small} prove that three-layer neural networks with ReLU (Rectified Linear Unit) activations can fit any arbitrary dataset when the number of hidden neurons equals the square root of the sample size, while  \cite{schmidt2020nonparametric} establish that sparsely connected deep ReLU networks can achieve optimal minimax estimation rate.

The majority of neural network research has been conducted outside of the field of statistics, resulting in a limited incorporation of statistically-based methodologies and a disparity in the terminology used \citep{hooker2021bridging}.
Neural networks, particularly MLPs, have emerged as strong alternatives to traditional statistical models, albeit they are often viewed as quite separate entities.
However, a number of researchers have illustrated the similarities between neural networks and statistical models. 
\cite{NN_as_Statmod_WarrenS} conducted a comprehensive comparison of various neural network models with traditional statistical approaches, showing that MLPs can be viewed as  generalisations of non-linear regression and discriminant models.
\cite{warner1996understanding} highlight the parallels between neural networks and statistical regression models in terms of notation and implementation, showing that neural networks act as a type of nonparametric regression model.
\cite{cheng1994neural} inform statistical readers about neural networks, highlighting similarities with statistical methodology.
More recently, \cite{ha2024fitting} utilise deep neural networks within both a generalised non-linear modelling framework (extending classical generalised linear models) and a non-linear survival model (extending the Cox model).
Furthermore, \citet{McInerney_2024} place neural networks within a statistical Gaussian likelihood-based framework, enabling input and architecture selection in a manner that is quite familiar within statistical modelling, but less so in neural networks research.
Synthesizing these views, it becomes evident that neural networks are a generalisation of statistical approaches whereby they automatically learn complex non-linear effects and interactions within their hidden layers; this is in contrast to the manual specification of interactions required in traditional statistical modelling, which frequently utilises simpler linear effects for practicality.

Given previous work comparing the mathematical structure of statistical models and neural networks, the question then arises as to how these methods compare with each other in practical applications.
Indeed, several authors have made strides at answering this over the years.
\cite{ANN_vs_Stats_medical} conducted a review and analysis of the literature comparing neural networks with linear, logistic, and Cox regression in the context of medical research.
\cite{STAT_Vs_NN_transport} provided a survey of applications in the transportation area, comparing neural networks with linear or logistic regression predominantly. 
\citet{ALAKA2018164} carried out a systematic review of bankruptcy prediction studies, comparing multiple discriminant analysis and logistic regression with six artificial intelligence techniques and proposing a framework to guide tool selection and the use of hybrid models for improved predictive performance.
\cite{shin2021machine} carried out a systematic literature review of machine learning techniques (random forests, decision and regression trees, support vector machines, and neural networks) versus conventional statistical models (logistic, Cox, and Poisson regression) for predicting heart failure readmission and mortality.
\cite{cho2021machine} carried out a systematic review of the same aforementioned machine learning methods and conventional statistical models for the prediction of myocardial infarction readmission and mortality. 
\cite{sun2022comparing} conducted a systematic review and meta-analysis of machine learning techniques, including random forests, support vector machines, neural networks, and decision trees, against statistical methods, linear and Cox regression, for predicting heart failure events.
\citet{NAZARETH2023119640} conducted a systematic literature review examining machine and deep learning techniques including neural networks across six financial domains, comparing their effectiveness to traditional statistical approaches (linear and logistic regression) in tasks such as stock prediction, portfolio management, and bankruptcy forecasting.
Similarly, \citet{ELAZAB2024124780} reviewed the use of machine and deep learning models including linear and logistic regression for Alzheimer’s disease diagnosis.
More generally, \cite{paliwal2009neural} carried out a review of comparative studies across various application domains, evaluating multilayered feedforward neural networks against linear regression, logistic regression, and discriminant analysis. 
Their review critically assessed the literature based on several key criteria, including sample size, adherence to the assumptions of the techniques, the method by which results are validated, the metrics used for comparison, and whether statistically significant differences were observed in the outcomes.

Despite the interest in comparing neural networks and statistical models, prior reviews have predominantly narrowed their view to particular subject areas, such as transportation and heart health, as mentioned above. 
Additionally, these reviews have focused on quite simple statistical models, such as linear regression, which inherently lack the capabilities to model non-linear effects that may be underlying in the data.
Therefore, such comparisons are not fair representations of what is possible in more advanced non-linear statistical models.
Specifically, Generalised Additive Models (GAMs) \citep{GAM, hastie1990generalized} are the most popular extension to linear regression, replacing linear terms with arbitrary non-linear functions.
The strength of GAMs is their ability to deal with non-linear relationships between predictor and response variables, while maintaining interpretability through their additive structure.

To date, there has been no systematic review of the literature comparing GAMs with neural networks in a wide range of application areas, and, therefore, our current paper closes this gap.
The purpose of the systematic review is to be a source of information on the types of studies that have carried out comparisons of these two model classes and the reported findings within these studies.
To move beyond just a summary of these papers, we have further extracted and analysed the values of the performance metrics provided within the papers, along with characteristics that can potentially explain observed differences (e.g., application area, year of study, sample size, number of predictor variables, and neural network complexity).
This dataset has been made available for further research at \url{https://github.com/jessicadoohan/gam-vs-nn-review}.

The remainder of the paper is structured as follows.
In Section~\ref{sec: Theoretical background}, we provide a theoretical background on GAMs and neural networks. 
Section~\ref{sec: Research questions} outlines the research questions guiding this systematic review. 
The methodology for conducting the review is detailed in Section~\ref{sec: Methods}.
The study selection process for included papers and a quality assessment of the final sample is provided in Section~\ref{sec: Study selection}.
Then Section~\ref{sec: Study characteristics} summarises the data attributes extracted from the reviewed literature.
Analysis and modelling of the performance metrics used to compare GAMs and neural networks is provided in Section~\ref{sec: Performance metric analysis}.
Finally, conclusions are presented in Section~\ref{sec: Conclusion}.

\section{Theoretical background}\label{sec: Theoretical background}
Before providing the details of our systematic review, we first provide some brief technical details on the models under consideration. 
Traditional statistical models and neural networks share the fundamental objective of modelling the relationship between a set of covariates and a response variable.
Let $Y \in \mathbb{R}$ denote the response variable and $X = (X_1, X_2, \dots, X_k)^\top \in \mathbb{R}^k$ the vector of $k$ covariates, where we restrict attention to tabular data. 
The general regression framework can be written as
\begin{equation}
    \mathbb{E}[Y \mid X = \mathbf{x}] = f(\mathbf{x}),
    \label{doohan:eqn1}
\end{equation}
where $\mathbb{E}[\cdot]$ is the expectation operator, and $f(\mathbf{x})$ is a function of the observed covariates $\mathbf{x} = (x_1, x_2, \dots, x_k)^\top$, encapsulating how the mean of $Y$ depends on $X$. 
The choice of this function defines the type of regression model employed and may take several well-known forms, each corresponding to a different class of models:

\begin{equation}
\label{eqn: reg functions}
    f(\mathbf{x}) = 
    \begin{cases}
        g_1(x_1) + g_2(x_2) + \dots + g_k(x_k) & \text{GAM}
        \\
        \textbf{NN}(x_1, x_2, \dots, x_k) &
        \text{Neural Network}, 
    \end{cases}
\end{equation}
where GAMs utilise smooth functions $g_j(x_j)$ ($j = 1, \ldots, k$) to model non-linear relationships and $\textbf{NN}(x_1, x_2, \dots, x_k)$ is a fully connected neural network (NN).

\subsection{Generalised additive models (GAMs)}
GAMs extend linear regression by allowing non-linear relationships between the predictor variables $\mathbf{x}$, and the response $Y$.
As seen in Equation~\ref{eqn: reg functions}, smooth functions $g(\cdot)$ replace the constant coefficients of linear regression, which increases flexibility.
Smooth functions can be approximated as a linear combination of a set of known basis functions, 
\begin{equation}
    g_j(x_j) = \sum_{p=1}^P c_p b_p(x_j),
\end{equation}
where $P$ is the number of specified functions, ($b_1(\cdot), \ldots, b_P(\cdot)$), controlling the maximum complexity of the smooth function, and the $c_p$ terms are associated constants to be estimated from data.
Common basis functions include cubic regression splines \citep{durrleman1989flexible}, thin-plate regression splines \citep{wood2003thin}, B-splines \citep{de1978practical}, and penalised B-splines (``P-splines'') \citep{eilers1996flexible}. 
The estimation of these spline functions, historically achieved via the backfitting algorithm \citep{GAM}, can now be conducted through penalised likelihood maximisation \citep{wood2011fast}. 
This approach integrates a roughness penalty directly into the fitting criterion to automatically prevent overfitting.
The theoretical and practical development of GAMs has been advanced substantially by Wood’s \texttt{mgcv} (\emph{Mixed GAM Computation Vehicle}) \citep{wood2017generalized, wood2012mgcv} package in the R programming language \citep{R_cite}.

A key advantage of GAMs is that their additive nature retains interpretability.
By plotting respective smooth functions, GAMs allow the visualisation of each covariate effect.
While GAMs have increased flexibility over linear regression, their additive nature, which aids in the interpretation, also limits flexibility.
Complex high-degree interactions can be difficult to estimate using splines and are often not specified in practice as a result.

\subsection{Neural networks}
\begin{figure}[H]
  \centering
  \includegraphics[width=0.65\linewidth]{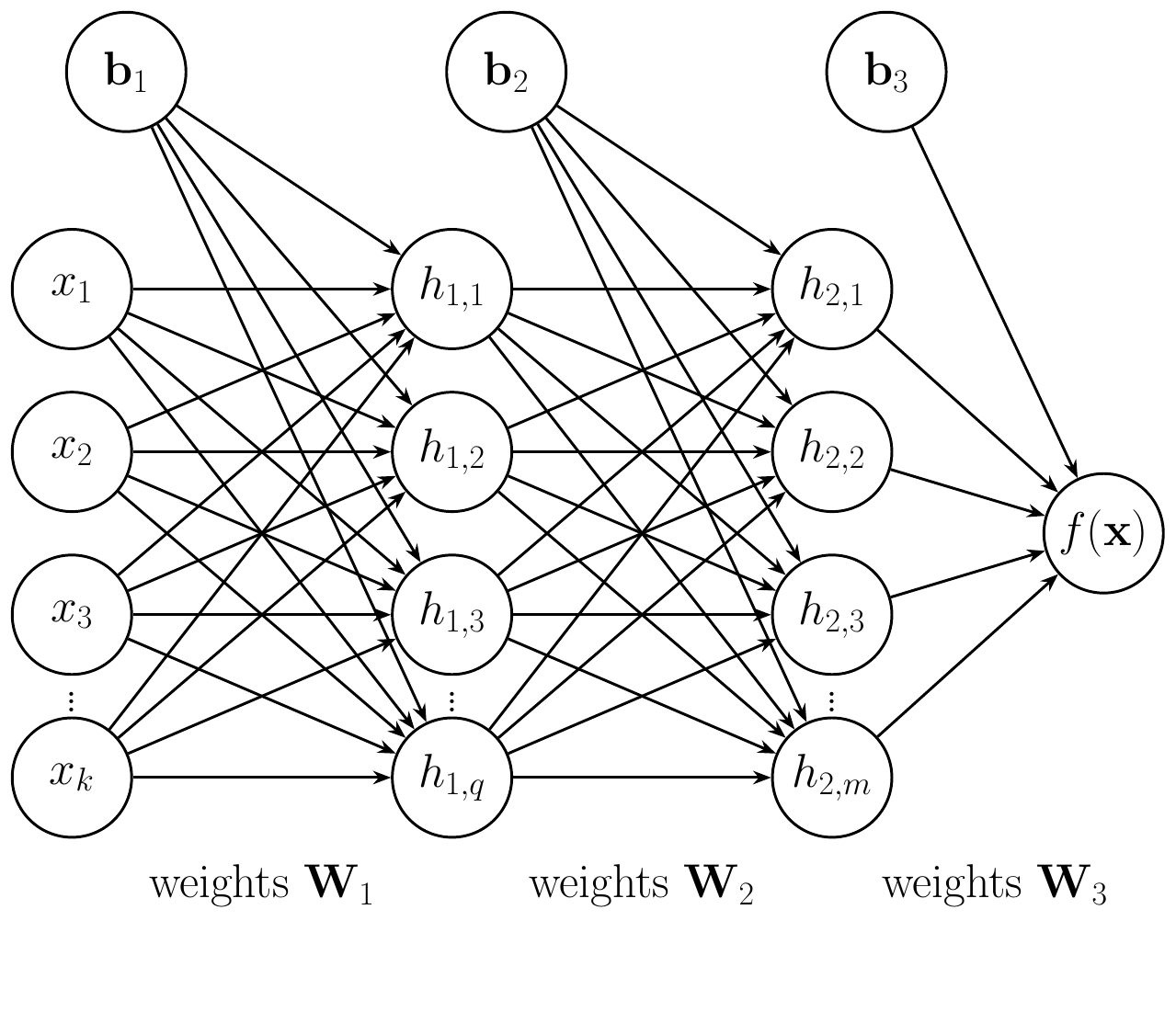}
  \caption{Multilayer perceptron neural network.}
  \label{fig: MLP_diagram}
\end{figure}

Neural networks are a class of models inspired by the structure of the human brain, where information flows through the interaction of neurons \citep{goodfellow2016deep}.
Neural networks offer a flexible approach to modelling $f(\mathbf{x})$ by approximating complex, non-linear relationships through layers of interconnected nodes.
A standard architecture is the MLP neural network.
This architecture passes the data through a series of weighted transformations and non-linear activation functions.
The default structure of a neural network is fully connected, essentially specifying all possible interactions (assuming a sufficiently complex neural network).
An MLP with $L$ layers can be described by the equation: 

\begin{equation}
    f(\mathbf{x}) = h_L(\mathbf{b}_L + \mathbf{W}_Lh_{L-1}(\dots  h_\ell(\mathbf{b}_\ell + \mathbf{W}_{\ell}h_{\ell-1}( \dots \mathbf{b}_2 + \mathbf{W}_2h_1(\mathbf{b}_1 + \mathbf{W}_1\mathbf{x}))))),
\end{equation}
where $\mathbf{b}_\ell$ is a bias vector,  $\mathbf{W}_\ell$ is a weight matrix, and $h_\ell(\cdot)$ is a non-linear activation function ($\ell = 1, \ldots, L$).
Figure~\ref{fig: MLP_diagram} illustrates the architecture of a multilayer perceptron neural network.
Neural networks are typically optimised via stochastic gradient descent and its variants \citep{bishop1995neural, bottou2010large}, which iteratively update parameters to minimise a loss function using small, random subsets of data (batches).

The structure of neural networks, while highly flexible, proves difficult to interpret, leading to the label ``black-box'' models, as understanding the effect of each individual predictor variable and their interactions is often lost in this complexity.
The flexibility of these models makes them strong predictors.
However, determining key architectural parameters such as the number of hidden layers, the number of nodes per layer, and other hyperparameters is not straightforward. 
Identifying the optimal configuration typically involves an extensive and computationally intensive search process.

    
\section{Research questions}\label{sec: Research questions}
Prior to conducting our systematic review of comparisons of GAMs and neural networks, we defined key research questions to structure this review and facilitate a targeted comparison between the two model classes.
Table~\ref{tab:research_questions} highlights our three central research questions along with the motivation behind each.

\begin{table}[H]
\caption{Research questions and motivations.}
    \centering
    \renewcommand{\arraystretch}{1.3}
    \begin{tabular}{c p{4.5cm} p{7cm}}
        \hline
        \textbf{RQ} & \textbf{Research questions} & \textbf{Motivation} \\
        \hline
        RQ1 & What are the key characteristics of papers carrying out comparisons between GAMs and NNs? & Identify domain areas, journals, performance metrics, model structures, and dataset attributes. Determine standard evaluation practices, ensure comparability across studies, and highlight potential research gaps. \\

        RQ2 & Which model type performs better? & Summarise author-reported conclusion on model performance and advance on this through independent analysis of performance metrics. \\

        RQ3 & What variables are associated with differences in performance of GAMs and NNs? & Identify factors such as sample size, number of predictors, and dataset characteristics that may be associated with model performance, and examine how the relative performance of these models has evolved over time. \\

        \hline
    \end{tabular}
    \label{tab:research_questions}
\end{table}

\section{Methods}\label{sec: Methods}
This review was reported following the Preferred
Reporting Items for Systematic Reviews and Meta-Analyses (PRISMA) guidelines \citep{PRISMA}.

\subsection{Eligibility criteria}
\label{section: eligibility criteria}
To be eligible for inclusion, papers had to be from peer-reviewed journals or conference proceedings, written in English, and have full-text availability.
Additionally, papers were required to include original experimental analysis of both GAMs and neural networks using the same dataset.
Papers without real-data analysis, such as review, discussion, and theoretical papers, were excluded.
Papers in which the values of performance metrics were unattainable or provided in-sample results only were excluded.
Duplicate papers and grey literature, including preprints and dissertations, were also excluded.

\subsection{Information sources }\label{sec: syst}
The search was conducted using Scopus, a large multidisciplinary database of peer-reviewed literature, including journal articles, books, and conference proceedings. 
The search was limited to titles and abstracts, ensuring that the identified papers were directly relevant to the comparison of GAMs and neural networks.
The final search was completed on November 20, 2024, with no restrictions on the earliest publication date, allowing for the inclusion of all relevant literature regardless of publication year.

\subsection{Search strategy}
To ensure a comprehensive and reproducible search, we formulated a search strategy incorporating synonyms for both GAMs and neural networks.
Boolean operators were applied, using ‘OR’ to incorporate variations of each term and ‘AND’ to ensure that both model types were jointly present in the retrieved studies.

The search string used was as follows:
(GAM OR ``Generalised additive model'' ) AND (``Neural network'' OR ``artificial-NN'' OR ``Artificial Neuron Network'' OR ``Multilayer perceptron'' OR MLP))  

The systematic search resulted in 445 documents that satisfied the search criteria.

\subsection{Selection process}
To determine whether a document would be included in the final systematic review, each study was evaluated against the set of predefined eligibility criteria described in 
\mbox{Section~\ref{section: eligibility criteria}.} Defining the inclusion criteria in advance provides consistency selecting each paper, maintains relevance of the selected papers to our research questions, reduces potential selection bias, and ensures reproducibility of our systematic review. To assess whether a given document satisfied these conditions, the entire document was reviewed. The first author conducted the screening of each paper to determine its eligibility for inclusion. 
In cases where inclusion was uncertain, the paper was discussed with the second and third authors, and a decision was made.
Twelve papers could not be retrieved and were therefore excluded. 
Given the total of 444 papers considered in our review, this exclusion is quite minor ($<\!3\%$). However, it may introduce a small non-response bias that should be kept in mind when interpreting the subsequent analysis. Applying our inclusion criteria to these 444 papers resulted in a total of 143 papers, from which comparisons on 430 datasets were included in our final systematic review (as many papers conduct comparisons on more than one dataset).

\subsection{Data collection process}
The information collected from the papers in our systematic review is summarised in Table~\ref{tab:dataitems}. 
The corresponding source column indicates whether the information was provided by the Scopus search engine, Scimago Journal Rank (SJR), or manually extracted from within the paper itself.
Domain area was extracted from Scopus, and then manually combined into higher-level categories by the authors.
When studies reported results from multiple neural networks or GAMs, for example, with varying hyperparameters or architectures within the same dataset, information was collected from the model with the best performance.
Note, from Table 2 (``Level'' column), that there are paper-level attributes (e.g., title and authors), and dataset-level attributes for each of the datasets analysed within a given paper (e.g., author-reported outcome and data type).

\begin{table}[H]\centering
\footnotesize
\caption{\label{tab:dataitems} Data attributes for which information was sought from papers.}
\medskip
{\footnotesize
\begin{tabular}{llll}
\toprule[0.09 em]
\textbf{Attribute}  & \textbf{Description} & \textbf{Source} & \textbf{Level} \\
\midrule
  Title & Paper title & Scopus & Paper \\ 
  Authors & Authors list &  Scopus & Paper\\ 
  Year & Publication year &  Scopus & Paper\\ 
  Source/journal title & Publication source & Scopus & Paper\\  
  Source/journal rank & SJR ranking  &  SJR & Paper\\
  Domain & Application area & Scopus/Manual & Paper\\
  Problem type & Regression or classification task & Manual & Dataset \\
  Validation & Method used, e.g., train-test & Manual & Dataset \\
  Sample size & Number of observations & Manual & Dataset \\  
  Predictors & Number of predictor variables & Manual & Dataset \\
  Neural network type & Neural network type, e.g., MLP & Manual & Dataset \\
  Neural network complexity & Layers and nodes count & Manual & Dataset\\ 
  Author-reported outcome & Author's conclusion on best model & Manual & Dataset\\ 
  Performance metrics & Performance metrics used and their values &  Manual & Dataset\\ 
\bottomrule[0.09 em]
\end{tabular}}
\end{table}

\section{Study selection}\label{sec: Study selection}
\begin{figure}[H]
  \centering
  \includegraphics[width=\linewidth]{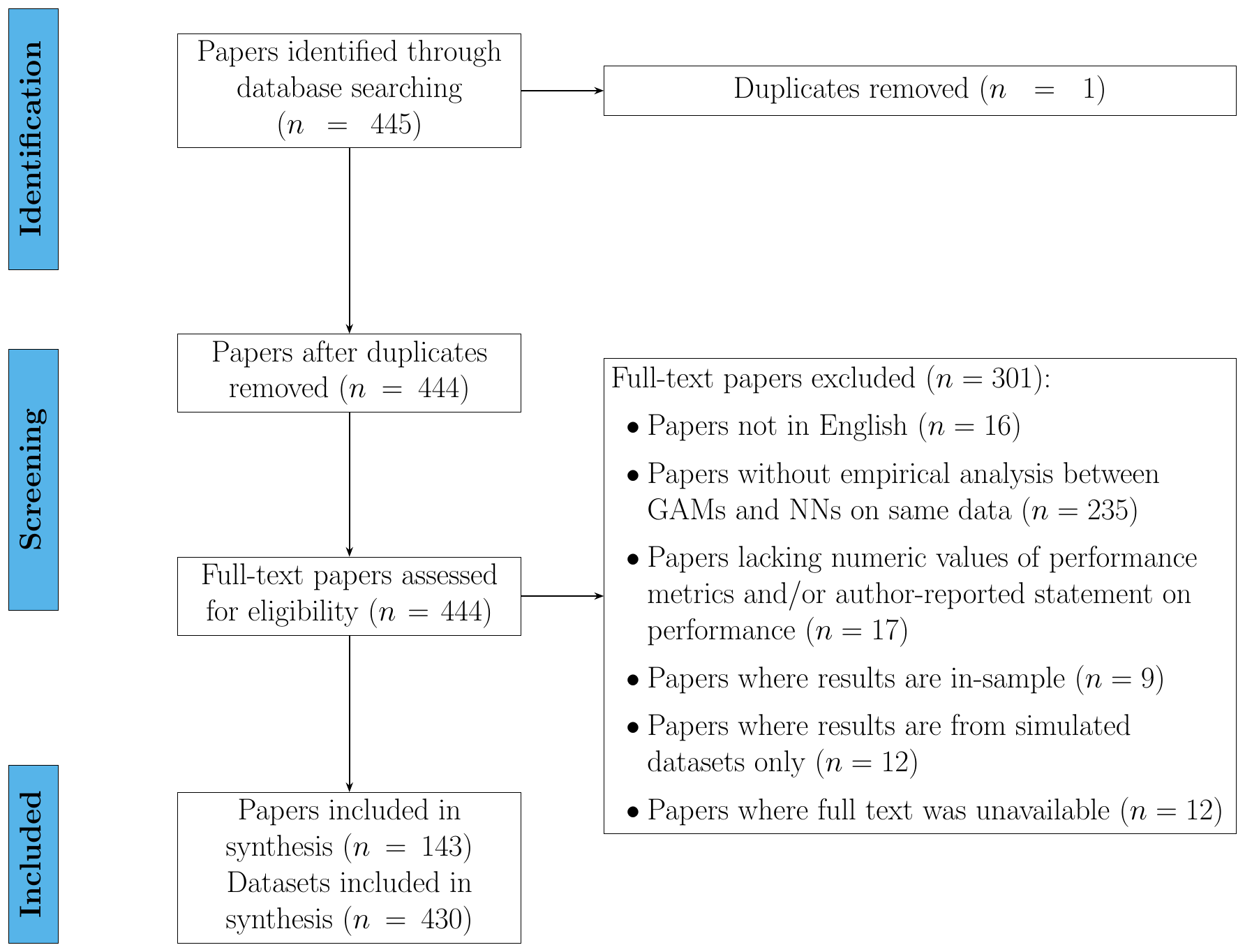}
  \caption{Flow diagram summarising the results of the paper identification, screening, and inclusion.}
  \label{fig: flow_diagram}
\end{figure}

The flow diagram in Figure~\ref{fig: flow_diagram} summarises the study selection process.
The search identified 445 papers, including one duplicate, which was removed.
Full-text of the remaining 444 papers was screened, of which 143 were included in the review.

\subsection{Quality assessment}\label{sec: Quality assessment}
Among the 15 attributes listed in Table~\ref{tab:dataitems}, sample size, number of predictors, and neural network complexity were the three that were not always reported in the papers, leading to missingness.
The frequency distribution of combinations of these variables and the quantity of datasets with each combination can be seen in Table~\ref{tab:missingness}.
Only 21\% of datasets report all three variables.
However, 79\% report at least two, and 95\% report at least one.
Unreported neural network complexity accounts for the majority of missingness, which is reported in only 35\% of datasets.

The level of missingness in the literature raises concerns about the level of transparency of reporting in the current literature comparing GAMs to neural networks.
Notably, this pattern is consistent with broader reporting practices observed across scientific fields.
For example, various systematic reviews have found high rates of unreported sample sizes (41\% in \cite{johnson2017rapid} and 12\% in \cite{arcaya2016research}).
In \cite{andaur2022completeness}, a systematic review of clinical prediction model reporting, found many instances of missing information, e.g., sample size (76\% missing), predictor variables included (86\% missing), model specification (95\% missing). Our data extraction process is therefore subject to the same limitations imposed by the original reporting quality observed within the literature.

Key methodological details, such as dataset characteristics and model complexity, are essential for evaluating the observed performance differences and for the reproducibility of the analyses carried out within these papers.
The lack of standardised reporting also suggests a need for using clear guidelines and policies to promote complete reporting of experimental details in publications of which researchers have made strides at formalising \citep{TriPod, mitchell2019model}.


\definecolor{myred}{HTML}{CC79A7}
\definecolor{mygreen}{HTML}{009E73}

\newcommand{\cmark}{\textcolor{mygreen}{\ding{52}}}
\newcommand{\xmark}{\textcolor{myred}{\ding{55}}}

\begin{table}[H]
\centering
\caption{\label{tab:missingness} Cross-tabulation of availability in collected attributes.}
\begin{tabular}{ccc|cr}
\toprule
\textbf{Sample size} & \textbf{Predictors} & \textbf{Complexity} & \textbf{Datasets} & \textbf{\%} \\
\midrule
\cmark     & \cmark     & \cmark     & \phantom{1}89 & 21\% \\
\cmark     & \cmark     & \xmark     & 189 & 44\% \\
\xmark     & \cmark     & \cmark     & \phantom{1}39 & 9\% \\
\cmark     & \xmark     & \cmark     & \phantom{1}23 & 5\% \\
\cmark     & \xmark     & \xmark     & \phantom{1}31 & 7\% \\
\xmark     & \cmark     & \xmark     & \phantom{1}39 & 9\% \\
\xmark     & \xmark     & \cmark     & \phantom{11}0 & 0\% \\
\xmark     & \xmark     & \xmark     & \phantom{1}20 & 5\% \\
\hline
332 & 356 & 151 & &\\
77\% & 83\% & 35 \% & &\\

\bottomrule
\end{tabular}
\end{table}

\section{Study characteristics}\label{sec: Study characteristics}
In this section we address RQ1, which asks about the key characteristics of papers comparing GAMs and NNs.
We summarise paper-level and dataset-level attributes extracted to characterise the scope and structure of the existing literature. Table \ref{tab:all papers} lists all 143 papers included in this systematic review, grouped by domain area.

\begin{table}[H]
\caption{\label{tab:all papers} Papers included in the systematic review.}
\footnotesize
\centering
{\footnotesize
\begin{tabular}{p{10cm}p{1cm}p{0.8cm}}
\toprule
 \textbf{Domain area} & \textbf{Papers} &\textbf{Datasets} \\ 
\midrule
\midrule
\textbf{Environmental \& Agricultural Sciences} & \textbf{66} & \textbf{199}  \\
\midrule
\midrule
\multicolumn{3}{p{\linewidth}}{  \cite{Aguilos2024}, \cite{Beaugrand2024}, \cite{Bonsoms20241777}, \cite{Fan2024}, \cite{Fu2024}, \cite{Gazis20242473}, \cite{Jung2024}, \cite{Lei2024}, \cite{Mouta202447}, \cite{Senaratna2024}, \cite{Yang2024}, \cite{Ghorbani20246096}, \cite{Adamova2023}, \cite{Boudreault2023}, \cite{DelSanto2023}, \cite{Li2023}, \cite{Yan2023339}, \cite{Benemann2022189}, \cite{Shafizadeh-Moghadam20221080}, \cite{Yadav2022194}, \cite{Evans2021}, \cite{Ma2021}, \cite{Martín2021289}, \cite{Mondal2021}, \cite{Phelps2021850}, \cite{Tian2021}, \cite{Zhang2021}, \cite{Dodangeh2020}, \cite{Dorich20201186}, \cite{Mandal2020}, \cite{Yadav2020129}, \cite{Mugo2020}, \cite{Nguyen2019}, \cite{Liu2018572}, \cite{Pourghasemi2018177}, \cite{Corona-Núñez201769}, \cite{Gao2017347}, \cite{Pôças2017177}, \cite{Ouarda20161553}, \cite{Shortridge20162611}, \cite{Vogel2016}, \cite{Durocher20151561}, \cite{Li2015312}, \cite{Lin20154088}, \cite{Xu201414}, \cite{Gutiérrez-Estrada201219}, \cite{Queirolo2012293}, \cite{Haywood20111174}, \cite{Palialexis2011165}, \cite{Aertsen20101119}, \cite{Gutiérrez-Estrada20101451}, \cite{Ogawa-Onishi20101728}, \cite{Tisseuil2010279}, \cite{Virkkala2010269}, \cite{Gutiérrez-Estrada2009116}, \cite{Lee2009367}, \cite{López-Darias20081027}, \cite{Araújo20051504}, \cite{Luoto2005299}, \cite{Megrey20051256}, \cite{Wang2005201}, \cite{Thuiller20031353}, \cite{Brosse20021033}, \cite{Brosse2000441}, \cite{Brosse19991293}, \cite{Shatar1999249} }
  \\ 
\midrule
\textbf{Engineering, Mathematics, \& Computer Science} & \textbf{35} &  \textbf{141} \\
\midrule
\midrule
\multicolumn{3}{p{\linewidth}}{  \cite{Chen202421429}, \cite{Kang2024}, \cite{Nascimento2024288}, \cite{Olca2024}, \cite{Pachauri2024}, \cite{Shojaeian20245371}, \cite{Berrisch2023}, \cite{Koh2023}, \cite{Ledmaoui20231004}, \cite{Montero2023222}, \cite{Razaq20233957}, \cite{Zhang20231}, \cite{Zhang2023}, \cite{Duan20223238}, \cite{Mestanza2022397}, \cite{Pachauri202224566}, \cite{Dudek2021}, \cite{Fang20213503}, \cite{Kamis2021}, \cite{Matsumoto2021}, \cite{Yang2021}, \cite{Daniel2020}, \cite{Park2020}, \cite{Ilseven2019}, \cite{Tsang20185804}, \cite{Ilseven20171}, \cite{Martín2016173}, \cite{Francisco-Fernández2015541}, \cite{Yang2015108}, \cite{Fukuda20131}, \cite{Aertsen2011929}, \cite{Berg2007129}, \cite{Schlink2006547}, \cite{Morlini2001169}, \cite{Faraway1998231}}
 \\ 
\midrule
\textbf{Health \& Biomedical Sciences} & \textbf{12}  & \textbf{25} \\
\midrule
\midrule
\multicolumn{3}{p{\linewidth}}{  \cite{Francisco2024}, \cite{Esmaeili2023423}, \cite{Xu2023}, \cite{Moslehi2022}, \cite{Shoko2022534}, \cite{Taylor2019}, \cite{Allard2017338}, \cite{Cengiz2012282}, \cite{Gregori2011277}, \cite{Gregori2009777}, \cite{Giachino20071657}, \cite{Goldfarb-Rumyantzev2002252}}
\\ 
\midrule
 \textbf{Other} & \textbf{30} & \textbf{65} \\
\midrule
\midrule
\multicolumn{3}{p{\linewidth}}{\cite{Pasha2024}, \cite{Roldán2024}, \cite{Szeląg2024}, \cite{Amara-Ouali20231272}, \cite{Balan20232278}, \cite{Bolleddula2023}, \cite{Cai20231432}, \cite{Krämer2023365}, \cite{Mathur2023289}, \cite{Sanii202268}, \cite{Zhou2022}, \cite{Dwivedi2021}, \cite{Fotiadis2021}, \cite{Frasier2021}, \cite{Li20211364}, \cite{Rengarajan2021}, \cite{Zhu2021}, \cite{Anglart20208433}, \cite{Zhu20201}, \cite{Zihan2020}, \cite{Wheeler2019}, \cite{Baquero2018}, \cite{Luan2018}, \cite{Davoudi2017568}, \cite{Nyhan2014305}, \cite{Liu20121431}, \cite{Ryser2009187}, \cite{Marmion20082241}, \cite{Heikkinen2007347}, \cite{Lee2002237} }
\\ 
\bottomrule
\end{tabular}}
\end{table}

\subsection{Paper-level characteristics}
\begin{figure}[H]
    \centering
    \includegraphics[width=\linewidth]{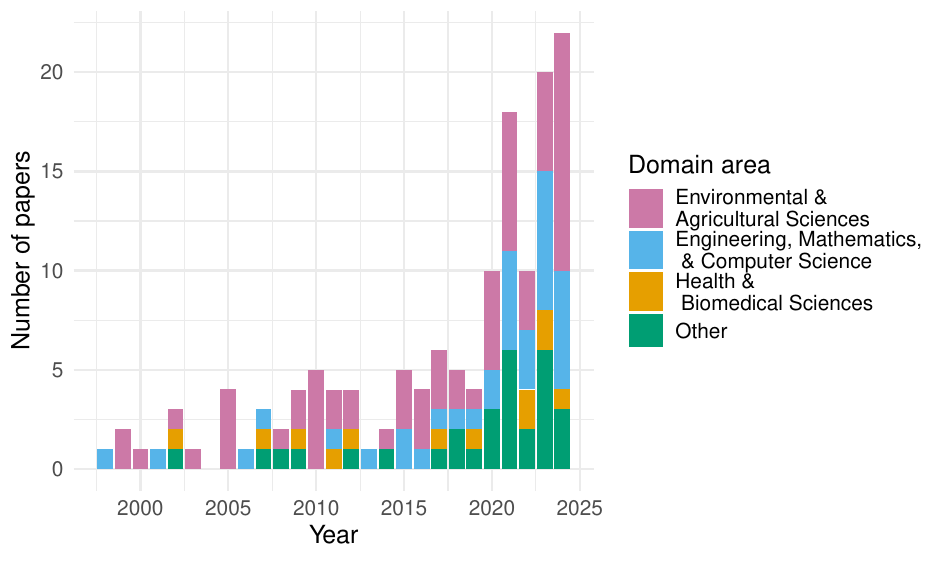}
    \caption{Number of papers plotted by year of publication, coloured by domain area.}
    \label{fig: areas by year}
\end{figure}

Year, publication source, publication rank, and domain area are four key paper-level characteristics that have been gathered from this review.
Figure~\ref{fig: areas by year} plots the number of papers by domain area through the years of publication.
There is a strong increasing trend in papers carrying out comparisons between GAMs and neural networks. 
The clear surge of activity in recent years emphasises the importance and timeliness of investigating this topic.

There are 121 distinct journals and conference proceedings from which the 143 papers originate.
Table~\ref{tab:journal rank} summarises the most frequently occurring publication sources, revealing that only six journals appear at least three times.
This indicates a broad range of research across multiple disciplines, reflecting the interdisciplinary nature of research comparing GAMs and neural networks.
To assess the academic impact of these publications, we analysed their rankings according to the Scimago Journal \& Country Rank (SJR) 2024. The distribution is as follows: 69\% of publications appear in Q1 (top quartile) journals,
13\% in Q2, 8\% in Q3, 3\% in Q4, and 8\% are currently unranked.
The majority of comparative studies between GAMs and neural networks are published in Q1 and Q2 sources, reinforcing the significance of this research area.

\begin{table}[H]
\caption{\label{tab:journal rank} Most frequent sources of the papers in this systematic review.}
\centering
{
\begin{tabular}{lcc}
  \toprule
\textbf{Journal title} & \textbf{No. papers} & \textbf{SJR best quartile}  \\ 
  \midrule
Ecological Modelling &   3 & Q1\\ 
  Science of the Total Environment &   3 & Q1  \\ 
    Fisheries Research &   3 & Q1 \\ 
  Environmental Modelling and Software &   3 & Q1 \\ 
  PLoS ONE &   3 & Q1  \\ 
  Remote Sensing &   3 & Q1  \\  

   \bottomrule
\end{tabular}}
\end{table}

High-level domain areas were assigned to each paper based on the areas assigned by Scimago to the source/journal in which the paper appeared.
The following four domain areas were used: Environmental and Agricultural Sciences, Engineering, Mathematics, and Computer Science, Health and Biomedical Sciences, and Other. 
Table~\ref{tab: paper areas} summarises domain areas by paper, dataset, and the average number of datasets per paper.

\begin{table}[H]
\caption{\label{tab: paper areas} Domain areas.}
\centering
{
\begin{tabular}{lllc}
  \toprule
\textbf{Domain area} & \textbf{Papers} & \textbf{Datasets} & \textbf{Avg. \# datasets} \\ 
  \midrule
  Env.\ \& Agri.\ Sci.\ (EAS) & 66\phantom{a}(46\%) & 199\phantom{a}(46\%) & 3 \\ 
  Eng., Math.\ \& Comp.\ Sci.\ (EMCS) & 35\phantom{a}(24\%) & 141\phantom{a}(33\%) & 4 \\   
  Health \& Biomed.\ Sci.\ (HBS) & 12\phantom{a}(\phantom{0}8\%) & \phantom{1}25\phantom{a}(\phantom{0}6\%) & 2 \\ 
  Other (OTH) & 30\phantom{a}(21\%) & \phantom{1}65\phantom{a}(15\%) & 2 \\ 
  \midrule
   Total & 143 & 430 & 3  \\
   \bottomrule
\end{tabular}}

\vspace{0.9em}
\parbox{0.9\linewidth}{\footnotesize
Abbreviations: EAS = Environmental \& Agricultural Sciences; 
EMCS = Engineering, Mathematics \& Computer Science; 
HBS = Health \& Biomedical Sciences; 
OTH = Other.
}
\end{table}

\subsection{Dataset-level characteristics}
\begin{figure}[H]
    \centering
    \includegraphics[width=\linewidth]{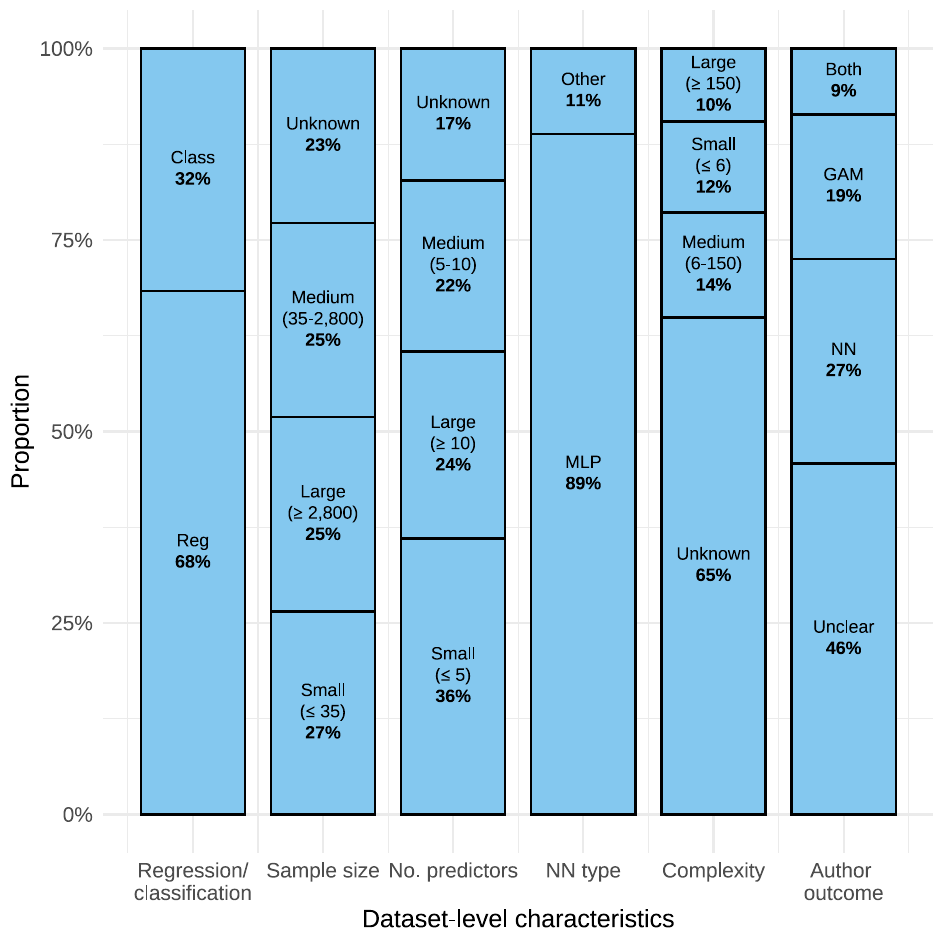}
    \caption{Proportion of datasets by key characteristics.
    Reg = regression; Class = classification. 
    Category labels (Small, Medium, Large, Unknown) refer to relative groupings based on sample size, number of predictors, and model complexity. }
    \label{fig: dataset level characteristics}
\end{figure}

Whether the dataset is a regression or classification task, sample size, number of predictors, type of neural network, neural network complexity, author-reported outcome, and performance metrics used are key dataset-level characteristics gathered from papers included in this review.
Figure~\ref{fig: dataset level characteristics} displays a summary of these characteristics.

Regression tasks accounted for 68\% of the datasets, while 32\% were classification.
Sample size, number of predictors, and neural network complexity are three key numerical variables extracted at the dataset level.
Descriptive statistics for each are presented in Table~\ref{tab: summary stats}.
These statistics provide insight into the range of the datasets included in the review, as well as the variability in model complexity.
As shown, sample sizes range from 35 to over 1.2 million observations, with a median of 1,400. 
The number of predictors varies widely, from as few as 2 to as many as 3,000, though the median and mean remain low at 7 and 22, respectively.
Neural network complexity, defined as the number of parameters, has a wide range from 2 to 120,000 parameters. 

MLP neural networks were the most common type, accounting for 89\% of datasets. 
All other neural networks were variations of this, and any one type accounted for less than $5\%$ of the datasets.
Neural network complexity was calculated as the total number of parameters, excluding those in the input layer. 
This exclusion was necessary because input nodes correspond to the number of predictors, which is already captured by the fact that we have collected the number of predictor variables as an attribute.
As mentioned, neural network complexity is among the data items frequently unreported. 
Only 35\% of datasets provided sufficient detail to compute the full parameter count. 
However, 64\% reported the number of hidden layers, which includes instances where layers were discussed but not the number of nodes in each needed to look at full complexity.
Depth of a neural network via the number of hidden layers is, in its own right, a feature of interest.
Among the datasets that did report the number of hidden layers, 70\% used a single hidden layer, 22\% had two hidden layers, and 7\% used between three and five layers. 
This distribution indicates that the majority of neural networks used in comparisons with GAMs were shallow architectures, typically with one or two hidden layers.

\begin{table}[H]
\caption{\label{tab: summary stats} Summary statistics of dataset-level numerical variables.}
\centering
\begin{tabular}{lccccc}
  \toprule
  \textbf{Characteristic} & \textbf{Min.} & \textbf{Median} & \textbf{Mean} & \textbf{Max.} & \textbf{Reported} \\ 
  \midrule
  Sample size & 35 & 1400 & 15,000 & 1,200,000 & 77\%\\ 
  Predictors & 2 & 7 & 22 & 3,000 & 83\% \\ 
  Complexity & 2 & 21 & 6600 & 120,000 & 35\% \\ 
   \bottomrule
\end{tabular}
 {\footnotesize\begin{tablenotes}\centering
	\item[]{Note: numbers have been displayed to two significant figures.}
	\end{tablenotes}}
\end{table}

A wide range of over 80 performance metrics were used across the reviewed datasets.
These included well-known metrics such as Root Mean Square Error (RMSE) and the coefficient of determination ($R^2$), as well as domain-specific or author-defined measures particular to specific applications.
Table~\ref{tab:metric freq} displays the ten most frequently used metrics, ranked by the number of datasets in which they appeared. 
RMSE was the most commonly reported, followed by $R^2$, and the Area Under the Receiver Operating Characteristic (AUC) curve.

\begin{table}[H]\centering
\caption{\label{tab:metric freq} Frequently used performance evaluation measures.}
\medskip
\begin{tabular}{lc}
\toprule[0.09 em]
\textbf{Performance metric} & \textbf{Datasets}  \\
\midrule
  RMSE & 229 (53\%) \\
  $R^2$ & 141 (33\%) \\ 
  AUC & 111 (26\%) \\ 
  Mean Absolute Error & \phantom{0}86 (20\%) \\  
  Cohen's Kappa & \phantom{0}40 (\phantom{0}9\%) \\  
  Bias & \phantom{0}29  (\phantom{0}7\%) \\
  True Skill Statistic & \phantom{0}22  (\phantom{0}5\%) \\
  \% Accuracy  & \phantom{0}18  (\phantom{0}4\%)\\
  $R^2$ adjusted & \phantom{0}17  (\phantom{0}4\%) \\ 
  Relative RMSE & \phantom{0}17  (\phantom{0}4\%) \\
\bottomrule[0.09 em]
\end{tabular}
\end{table}

Looking at author-reported outcomes gives us an insight into RQ2, which investigates which model is better overall.
Neural networks were declared better than GAMs in 27\% of datasets.
This is more than the 19\% of datasets in which GAMs were reported as better than neural networks.
In 9\% of datasets, both model types were found equivalent.
However, author-reported outcomes provide an incomplete and often inconsistent comparison between GAMs and neural networks.
Differences can arise in how outcomes are classified from author to author.
Additionally, 46\% of author outcomes are undefined and hence recorded as ``unclear'', often due to a focus on an alternative model in that application.
A way to address these limitations is to directly analyse the performance metric values reported for both GAMs and neural networks in each instance.
This allows for a more objective comparison and quantifies the magnitude of performance differences, which is not possible when looking at the qualitative author-reported outcomes alone; it also provides a firmer, more quantitative answer to RQ2.

\section{Performance metric analysis}\label{sec: Performance metric analysis}
To quantitatively synthesize the comparative performance of GAMs and neural networks across the reviewed literature, we create a statistical model for the reported performance metric values directly.
This analysis examines how the relative performance, measured by the most frequently used metrics, is associated with key study characteristics such as sample size, domain area, and model complexity. 
This will address RQ2 and RQ3. 
RQ3 examines whether or not available data or modelling characteristics can explain observed differences in model performance. To answer these, we model the reported performance metrics from the 430 datasets in terms of the available characteristics.
Given the hierarchical structure of the data we have collected, where datasets are nested within papers, a mixed-effects modelling approach was employed to account for intra-paper dependence.

\subsection{Methods}
\label{sec:LMM methods}
Mixed-effects models extend linear regression by enabling the analysis of  multiple sources of variation in the data, rather than assuming a single source. 
These models allow the response variable to be modelled through both fixed and random effects.
The fixed effects estimate the average response across the entire population, similar to traditional regression, while the random effects (e.g., the influence of individual papers) capture variability across papers and allow for the estimation of paper-specific responses. 
In this analysis, many papers contain multiple datasets. 
This introduces a potential dependency (i.e., correlation) in the results, which are likely influenced by the same group of authors applying similar modelling strategies to the datasets they consider. 
Furthermore, the datasets themselves may be similar, coming from the same (potentially niche) application area or research goal, reducing their independence.
Hence, results derived from datasets within the same paper are typically correlated, and this intra-paper dependence must be considered in the analysis.
Mixed-effects models capture the correlation between observations in the same group, i.e., datasets within the same paper, known as intraclass correlation (ICC):
\begin{equation}
    \text{ICC} = \frac{\sigma^2_{paper}}{\sigma^2_{paper} + \sigma^2_{residual}},
\end{equation}
where $\sigma^2_{paper}$ is the variation across papers and $\sigma^2_{residual}$ is the variation within each paper after accounting for fixed effects and the group intercept.
ICC can be interpreted as the correlation between results in datasets from the same paper and also as the proportion of total variance that can be explained by the paper effect.
A large ICC indicates the appropriateness of using mixed-effects models compared to standard regression, as the paper effect is strong.

Linear mixed-effects models were fitted using the \texttt{lme4} package \citep{lme4_cite} in R programming language \citep{R_cite}.
We included a paper-level random intercept effect to account for correlated results within a given paper, along with the following fixed effects: publication year, domain area, sample size, number of predictors, and neural network complexity.
Pairwise interactions with year of publication were also included to account for trends over time.

\subsection{Response and predictor variables}
As shown in Table~\ref{tab:metric freq}, RMSE, $R^2$, and AUC are, by far, the most commonly-used performance metrics, accounting for 53\%, 33\%, and 26\% of datasets respectively and collectively covering 89\% of all datasets.
RMSE and $R^2$ are common performance metrics used in regression, accounting for 76\% and 48\% of regression datasets respectively, and 91\% collectively.
AUC, a standard metric for classification models, was used in 82\% of the classification datasets. 
These three metrics were selected for our performance metric analysis due to their prevalence and large coverage of datasets.

\begin{figure}[H]
    \centering
    \includegraphics[width=\linewidth]{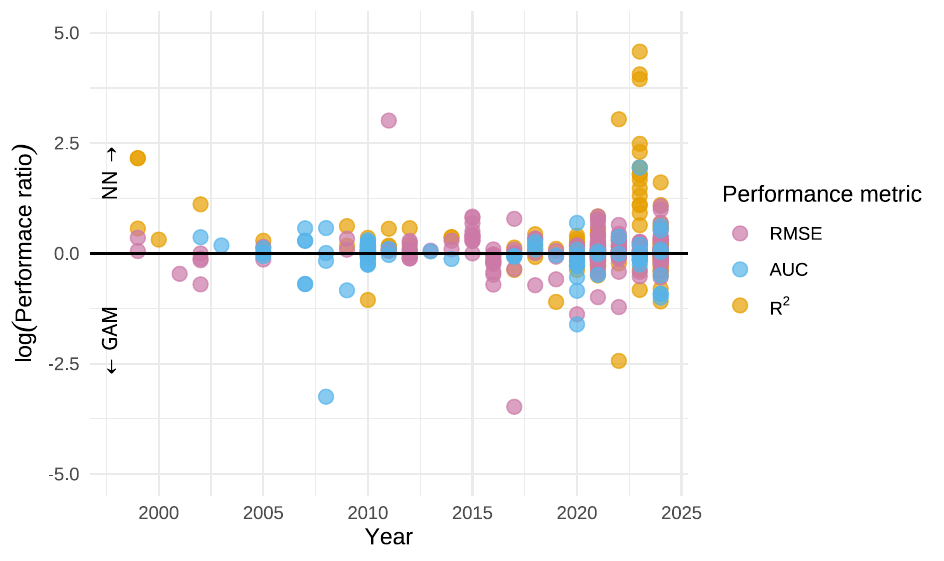}
    \caption{Log performance ratios plotted by year, where greater than zero signifies neural networks are outperforming GAMs and below zero indicates GAM superiority, as shown by arrows.}
    \label{fig: raw log ratio points}
\end{figure}

The log-ratio of each metric is computed as:
\begin{equation}
    \log(\mathrm{metric}_{\mathrm{ratio}}) = 
    \begin{cases}
        \log(\mathrm{RMSE}_{\mathrm{GAM}}/\mathrm{RMSE}_{\mathrm{NN}}) & \text{for RMSE} \\
        \log((1 -R^2_{\mathrm{GAM}})/ (1 - R^2_{\mathrm{NN}})) & \text{for } R^2 \\
        \log( (1 - \mathrm{AUC}_{\mathrm{GAM}})/ (1 - \mathrm{AUC}_{\mathrm{NN}})) &  \text{for AUC},
    \end{cases}
\end{equation}
where positive values indicate superior neural network performance for that metric, negative values signify GAM performance was better, and zero indicates model equivalence.
For AUC and $R^2$, they are first converted to losses, in which 0 indicates optimal performance, by subtracting them from 1.
This allows for all three to represent loss ratios.
The ratio of performance metrics provides a relative performance measure that is invariant to scale. 
This is necessary here, as we are comparing results across many different datasets with substantial variation in absolute performance; different datasets lie on different scales, with differing levels of achievable performance given the inherent noise/uncertainty in the problem at hand.
Moreover, we have taken the logarithm to place the response on the real line (so that it is suitable for linear modelling) and to address some skewness.
Figure~\ref{fig: raw log ratio points} displays the log performance ratios for datasets using RMSE, AUC, and $R^2$; points are coloured based on the performance metric used.
The horizontal line at zero indicates where GAMs and neural networks are equivalent; points above the line correspond to datasets where neural networks outperform GAMs, while points below are where GAMs achieve superior performance.
Overall, we can see that there is a high level of variability in results, with no firm evidence in favour of neural networks or additive models (as the points are generally centred around zero), and no strong trends over time. Thus, while neither model class appears to be systematically superior across the studies (addressing RQ2), this high-level overview of results does not account for study characteristics (other than the year) or the random effects of papers. These additional aspects are the focus of RQ3, and will be considered throughout the remainder of this paper.

The following study characteristics (i.e., paper and dataset characteristics) were considered in the meta-analysis as fixed effects: year of publication (\texttt{Year}), domain area (\texttt{Area}), sample size (\texttt{Sample\_Size}), number of predictors (\texttt{Predictors}), and neural network complexity (\texttt{Complexity}).
Paper ID (\texttt{Paper\_ID}) is used as a random effect to account for the dependence between datasets coming from the same papers.
To retain as much information as possible, the numerical variables with missing values (described in Section \ref{sec: Quality assessment}) were converted to categorical variables. Categories were formed by splitting the numerical variables by their observed tertiles, along with a further ``Unknown'' category for those with missing values, which resulted in four levels. This approach enabled the inclusion of incomplete records and allowed assessment of whether the presence of missingness itself was associated with performance outcomes. A detailed overview of all variables included in our meta analysis model for RQ3 is provided in Table \ref{tab:variables_summary}.

\begin{table}[H]
\centering
\caption{Summary of response variable, predictor variables, transformations, and associated boundary conditions used in the meta analysis model.}
\label{tab:variables_summary}
{\footnotesize
\renewcommand{\arraystretch}{1.25}
\begin{tabular}{p{4.5cm} p{10cm}}
\hline
\textbf{Component \& Type} & \textbf{Description \& boundary conditions} \\
\hline

\multicolumn{2}{l}{\textbf{Response variables}} \\

\mbox{Log performance ratio} \hspace{4cm}
Continuous (real number) &
Defined as\vspace{-0.2cm}
\begin{itemize}
\item$\log(\mathrm{RMSE}_\mathrm{GAM}/\mathrm{RMSE}_\mathrm{NN})$
\item$\log\{(1-R^2_\mathrm{GAM})/(1-R^2_\mathrm{NN})\}$
\item $\log\{(1-\mathrm{AUC}_\mathrm{GAM})/(1-\mathrm{AUC}_\mathrm{NN})\}$
\end{itemize}
Note that $R^2$ and AUC were first converted to losses ($1-R^2$ and $1 - \mathrm{AUC}$) to align with RMSE, and the log transformation places the responses on the real line such that positive values imply that NNs are better and negative values imply that GAMs are better.
 \\

\hline
\multicolumn{2}{l}{\textbf{Predictor variables (fixed effects)}} \\

\mbox{\texttt{Year}} \hspace{4cm}
Numerical (integer)  & Year of study (1998 to 2024) centred by subtracting the middle year, 2011 (leading to the range $-13$ to $13$). \\[0.1cm]

\mbox{\texttt{Area}} \hspace{4cm}
Categorical (4 levels) &
Domain area classified from Scopus categories as: Environmental and Agricultural Sciences; Engineering, Mathematics and Computer Science; Health and Biomedical Sciences; and Other. \\[0.1cm]

\mbox{\texttt{Sample\_Size}} \hspace{4cm}
Categorical (4 levels) &
 Number of observations in dataset: Small $\le 350$, Medium ($350$--$2800$), Large $\ge 2800$, and Unknown. \\[0.1cm]

\mbox{\texttt{Predictors}} \hspace{4cm}
Categorical (4 levels) & Number of predictors in dataset:
Small $\le 5$, Medium ($5$--$10$), Large $\ge 10$, and Unknown. \\[0.1cm]

\mbox{\texttt{Complexity}} \hspace{4cm}
Categorical (4 levels) &
Number of neural network parameters (excluding input layer): Small $\le 6$, Medium ($6$--$150$), Large $\ge 150$, and Unknown. \\[0.1cm]

\hline
\multicolumn{2}{l}{\textbf{Random effect}} \\

\mbox{\texttt{Paper\_ID}} \hspace{4cm} 
Categorical (143 levels) &
The ID of the paper, which accounts for correlation of multiple datasets originating from that paper. \\

\hline
\multicolumn{2}{l}{\textbf{Additional modelling details}} \\

Pairwise interactions &
Interactions between \texttt{Year} and the other predictors were fitted to investigate whether effects changed over time. \\[0.1cm]

Reference levels 
  &
For \texttt{Year}: 2011 (centred to be 0). For categorical variables \texttt{Sample\_Size}, \texttt{Predictors}, and \texttt{Complexity}: ``Small''.\\

\hline
\end{tabular}
}
\end{table}

\begin{figure}[H]
  \centering
  \includegraphics[width=0.8\textwidth]{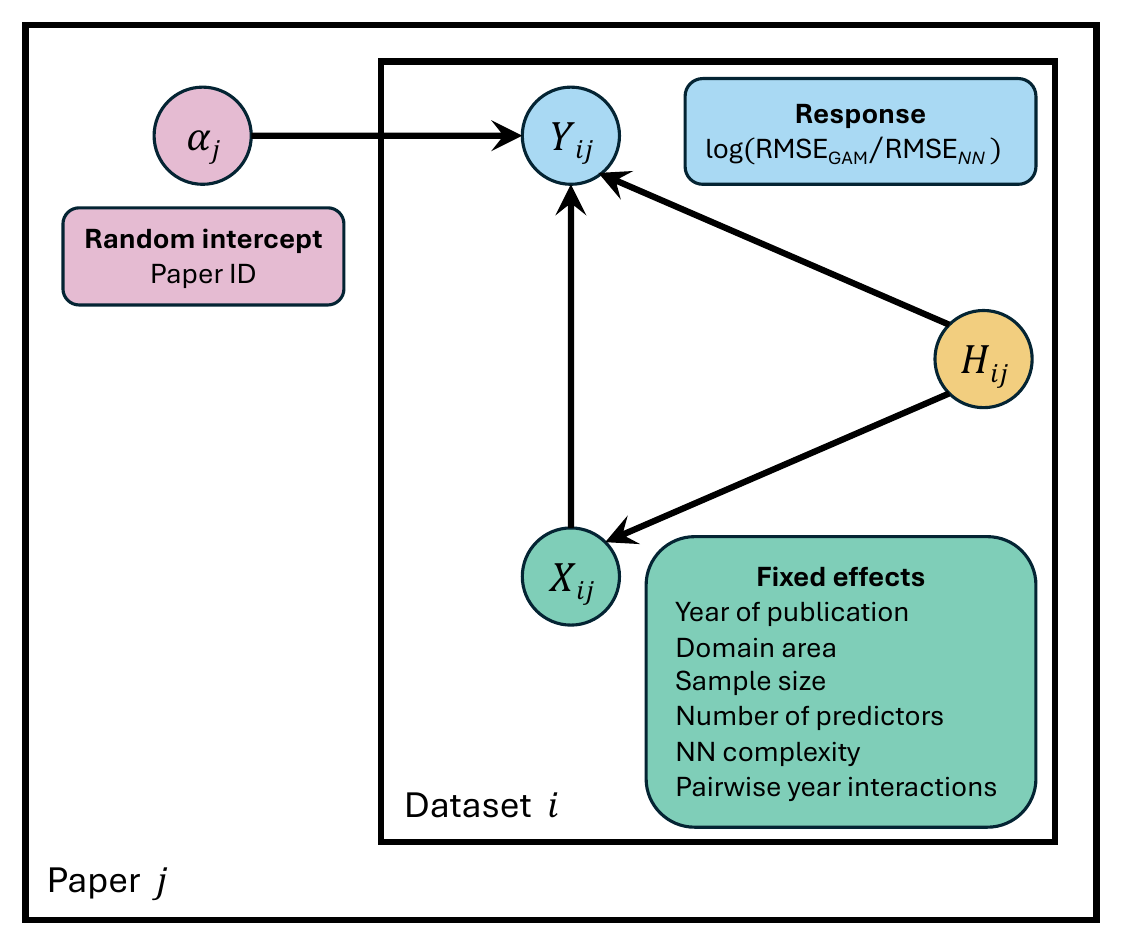} 
  \caption{Linear mixed-effects model diagram where $Y_{ij}$ is the response for the RMSE model case. For the $R^2$ model the response is $\log\{(1-R^2_\mathrm{GAM})/(1-R^2_\mathrm{NN})\}$ and AUC model response is $\log\{(1-\mathrm{AUC}_\mathrm{GAM})/(1-\mathrm{AUC}_\mathrm{NN})\}$. The random intercept is denoted by $\alpha_j$, $X_{ij}$ are the set of fixed-effects, and $H_{ij}$ are any potential latent variables.}
  \label{fig: LMM_diagram}
\end{figure}

We consider the linear mixed–effects model with random
intercepts:
\begin{equation}
  Y_{ij}
  = \beta_0 + \alpha_j + X_{ij}\beta + \varepsilon_{ij},
  \qquad
  \alpha_j \sim N(0,\sigma_{\alpha}^2), \quad
  \varepsilon_{ij} \sim N(0,\sigma^2),
  \label{eq:random-intercept-model}
\end{equation}
where $i$ indicates datasets within paper $j$.
Here $\beta_0$ and $\beta$ are fixed effects, while $\alpha_j$ represents the paper‐specific random intercept capturing paper-level heterogeneity.
The residual term $\varepsilon_{ij}$ represents dataset-level noise, assumed independent given the fixed and random effects. With the notation of \eqref{eq:random-intercept-model} in place, note that the intraclass correlation, described Section \ref{sec:LMM methods}, is given by $\mathrm{ICC} = \sigma_{\alpha}^2/(\sigma_{\alpha}^2+\sigma^2)$.
Figure~\ref{fig: LMM_diagram} provides a graphical representation of the random-intercept model in~\eqref{eq:random-intercept-model}.  
The outer box corresponds to a paper $j$, and the inner box represents a dataset $i$ within that paper.  
The random intercept $\alpha_j$ is a paper-level effect that influences all outcomes $Y_{ij}$ in that paper.  
The arrow from $\alpha_j$ to $Y_{ij}$ is the shared paper-specific shift in the mean response.  
The relationship between $X_{ij}$, $Y_{ij}$, and any unobserved latent variables $H_{ij}$ is shown inside the dataset box.

\subsubsection{Unrecorded variables} \label{sec: unrecorded variables}
Several additional potentially important dataset-level variables were not consistently reported within the reviewed literature, including data preprocessing steps, optimisation algorithms and settings, software environments, neural network hyperparameter tuning procedures, activation functions, and GAM details (e.g., spline type, degrees of freedom, knot placement, and smoothness penalty). 
These variables can influence model performance and therefore represent potential confounding variables in comparative analyses. 
However, because most papers did not report these in sufficient detail, they could not be incorporated as variables in this analysis.
The absence of such information likely introduces additional heterogeneity that cannot be modelled directly, but may be captured to some extent by the paper random effects included in our meta analysis model.

This limitation reflects a broader reporting issue common within machine learning research.
There is an absence of standardised reporting practices for model development and evaluation. 
As highlighted in \cite{mitchell2019model}, there is a growing recognition that inadequate documentation obscures methodological choices and hinders reproducibility of results.
Our systematic review strongly reinforces this previous research as we find that unreported information is prevelant across a large number of papers, from different fields, published over a 25-year period. Thus, there is a need for more rigorous reporting protocols even in simpler tabular dataset modelling.

Due to the unreported variables, the effects in the following sections are interpreted as associations rather than as causal mechanisms. Moreover, missing information is a common feature of systematic reviews in general as discussed previously in Section \ref{sec: Quality assessment}. In any case, it is of interest to explore the associations between model performance and key variables (e.g., year of study, discipline, data dimensions) per RQ3 within this systematic review, which is the first review of papers comparing neural networks and GAMs.

\subsection{RMSE metric analysis}\label{sec: RMSE analysis}
The RMSE analysis included 229 dataset observations from 80 papers. 
After removing two outliers, 227 dataset observations, from 78 papers remained. 
Including covariates in the analysis allows us to explain more of the variation in RMSE ratios across datasets, and to investigate how aspects of the datasets and models are associated with the relative performance of neural networks and GAMs. The RMSE model estimates and standard errors (SE) are presented in Table~\ref{tab:full_model}. 
(Models for AUC and $R^2$ are also shown, but discussed in Section~\ref{sec: AUC metric analysis} and Section~\ref{sec: R2 metric analysis} respectively.)
The model included all fixed-effects predictor variables and their pairwise interactions with time (\texttt{Year}) as follows:
\begin{equation}
\log(\mathrm{RMSE}_{\mathrm{ratio}}) \sim \beta_0 + \texttt{Year} + \texttt{Area} + \dots + (\texttt{Year} \times \texttt{Complexity}) + ( 1 | \texttt{Paper\_ID} ), 
\end{equation}
where $( 1 | \texttt{Paper\_ID} )$ represents a random intercept capturing variation among datasets.
Note that modelling the ratio in this manner is a special case of a slightly more general model discussed in~\ref{app: model alternative}.

\begin{table}[H]
\centering
\caption{Estimates from mixed-effects models.} 
\label{tab:full_model}
\begin{threeparttable}
{\footnotesize
\begin{tabular}{lccc}
   \toprule
   & \textbf{RMSE} & \textbf{AUC} & $\boldsymbol{R}^{\boldsymbol{2}}$\\ 
  \textbf{Term} & \textbf{Estimate (SE)} &  \textbf{Estimate (SE)} & \textbf{Estimate (SE)} \\ 
  \midrule
(Intercept) & $\phantom{-}$0.11 (0.28)\phantom{$^*$}& $-1.84$ (1.61)\phantom{$^*$} & $-0.70$ (1.11)\phantom{$^*$} \\ 

\texttt{Year} & $\phantom{-}0.02$ (0.03)\phantom{$^*$} & $\phantom{-}0.35$ (0.21)\phantom{$^*$} & $\phantom{-}0.11$ (0.11)\phantom{$^*$} \\ 
 
\texttt{Area$_\texttt{Env/Agri}$} & $\phantom{-}0.27$ (0.30)\phantom{$^*$} & $-0.72$ (1.54)\phantom{$^*$} & $\phantom{-}0.79$ (0.98)\phantom{$^*$} \\
\texttt{Area$_\texttt{Health/Bio}$} & $-0.32$ (0.45)\phantom{$^*$} & $\phantom{-}1.33$ (1.28)\phantom{$^*$} & $\phantom{-}0.22$ (1.04)\phantom{$^*$} \\
\texttt{Area$_\texttt{Other}$} & $-0.12$ (0.42)\phantom{$^*$} & $-0.42$ (1.35)\phantom{$^*$} & $-0.07$ (1.26)\phantom{$^*$} \\

\texttt{Sample\_Size$_{medium}$} & $\phantom{-}0.33$ (0.33)\phantom{$^*$} & $\phantom{-}3.59$ (0.27)$^*$ & $\phantom{-}0.52$ (0.59)\phantom{$^*$} \\ 
\texttt{Sample\_Size$_{large}$} & $\phantom{-}0.86$ (0.44)\phantom{$^*$}  & $\phantom{-}2.76$ (0.24)$^*$ & $\phantom{-}1.03$ (1.34)\phantom{$^*$}  \\ 
\texttt{Sample\_Size$_{unknown}$} & $\phantom{-}1.25$ $(0.35)^*$ & $\phantom{-}2.55$ (0.76)$^*$ & $\phantom{-}0.25$ (0.86)\phantom{$^*$} \\ 
 
\texttt{Predictors$_{medium}$} & $\phantom{-}0.26$ (0.23)\phantom{$^*$} & $-$0.08 (0.24)\phantom{$^*$} & $\phantom{-}$0.34 (0.43)\phantom{$^*$} \\ 
\texttt{Predictors$_{large}$} & $\phantom{-}$0.35 (0.34)\phantom{$^*$} &  $\phantom{-}$2.19 (0.70)$^*$ & $\phantom{-}$0.50 (0.82)\phantom{$^*$}  \\ 
\texttt{Predictors$_{unknown}$} & $\phantom{-}$0.11 (0.38)\phantom{$^*$} & $\phantom{-}$0.89 (1.04)\phantom{$^*$} & $\phantom{-}$0.25 (0.78)\phantom{$^*$} \\ 
 
\texttt{Complexity$_{medium}$} & $-$1.77 $(0.38)^*$ &  & $-$0.62 (0.62)\phantom{$^*$} \\
\texttt{Complexity$_{large}$} & $-$1.48 $(0.51)^*$ &  & $-$0.26 (0.91)\phantom{$^*$} \\ 
\texttt{Complexity$_{unknown}$} & $-$0.55 (0.29)\phantom{$^*$} & $-$1.07 (1.02)\phantom{$^*$} & $\phantom{-}$0.19 (0.52)\phantom{$^*$} \\
 
\texttt{Year} × \texttt{Area$_\texttt{Env/Agri}$} & $-$0.02 (0.03)\phantom{$^*$} &  $\phantom{-}$0.02 (0.17)\phantom{$^*$} & $-$0.09 (0.09)\phantom{$^*$} \\
\texttt{Year} × \texttt{Area$_\texttt{Health/Bio}$} & $\phantom{-}$0.01 (0.05)\phantom{$^*$} &  $-$0.04 (0.14)\phantom{$^*$} &  \\ 
\texttt{Year} × \texttt{Area$_\texttt{Other}$} & $\phantom{-}$0.02 (0.04)\phantom{$^*$} & $\phantom{-}$0.05 (0.15)\phantom{$^*$} & $\phantom{-}$0.03 (0.12)\phantom{$^*$} \\ 
 
\texttt{Year} × \texttt{Sample\_Size$_{medium}$} & $-$0.02 (0.03)\phantom{$^*$} & $-$0.32 (0.02)$^*$ & $-$0.02 (0.06)\phantom{$^*$} \\ 
\texttt{Year} × \texttt{Sample\_Size$_{large}$} & $-$0.06 (0.04)\phantom{$^*$} & $-$0.29 (0.05)$^*$  & $-$0.07 (0.12)\phantom{$^*$} \\
\texttt{Year} × \texttt{Sample\_Size$_{unknown}$} & $-$0.10 $(0.03)^*$ &  $-$0.21 (0.09)$^*$ & $\phantom{-}$0.00 (0.08)\phantom{$^*$} \\ 

\texttt{Year} × \texttt{Predictors$_{medium}$} & $-$0.03 (0.02)\phantom{$^*$} & $-$0.03 (0.06)\phantom{$^*$} & $-$0.03 (0.05)\phantom{$^*$} \\ 
\texttt{Year} × \texttt{Predictors$_{large}$} & $-$0.03 (0.04)\phantom{$^*$} & $-$0.25 (0.09)$^*$  & $-$0.06 (0.08)\phantom{$^*$} \\ 
\texttt{Year} × \texttt{Predictors$_{unknown}$} & $-$0.01 (0.04)\phantom{$^*$} & $-$0.15 (0.13)\phantom{$^*$} & $\phantom{-}$0.00 (0.08)\phantom{$^*$} \\

\texttt{Year} × \texttt{Complexity$_{medium}$} & $\phantom{-}$0.11 (0.04)$^*$ & & $\phantom{-}$0.04 (0.07)\phantom{$^*$} \\ 
\texttt{Year} × \texttt{Complexity$_{large}$} & $\phantom{-}$0.10 (0.05)$^*$ & & $\phantom{-}$0.01 (0.08)\phantom{$^*$} \\ 
\texttt{Year} × \texttt{Complexity$_{unknown}$} & $\phantom{-}$0.01 (0.03)\phantom{$^*$} &  $-$0.06 (0.13)\phantom{$^*$} & $-$0.09 (0.05)\phantom{$^*$} \\
  
$\sigma^2_{paper}$ & $\phantom{-}$0.27 \phantom{(0.00)$^*$} & $\phantom{-}$1.06 \phantom{(0.00)$^*$} & $\phantom{-}$0.79 \phantom{(0.00)$^*$}\\
$\sigma^2_{residual}$ & $\phantom{-}$0.08  \phantom{(0.00)$^*$} & $\phantom{-}$0.03 \phantom{(0.00)$^*$} & $\phantom{-}$0.23 \phantom{(0.00)$^*$}\\
ICC &  $\phantom{-}$0.76  \phantom{(0.00)$^*$} & $\phantom{-}$0.97 \phantom{(0.00)$^*$} & $\phantom{-}$0.78 \phantom{(0.00)$^*$}\\
   \bottomrule
\end{tabular}}
 {\footnotesize\begin{tablenotes}
	\item[]{($^{*}$) indicates statistical significance at the 5\% level.\\[-0.4cm]}
	\end{tablenotes}}
\end{threeparttable}
\end{table}

The intercept for this model is positive, however, not statistically significantly different from zero.
This indicates that, at the reference levels for covariates, there is no evidence of a distinction between the performance of GAMs and neural networks (as already noted in Figure~\ref{fig: raw log ratio points}).
Neural network complexity and its time interaction are statistically significant.
Specifically, medium and large complexity effects are both negative and significant, indicating that, in 2011, more complex neural networks performed worse than GAMs compared to smaller neural networks. 
The complexity-year interactions for medium and large neural networks are positive and significant; this indicates that larger neural networks are associated with improved performance over time relative to small neural networks.

\begin{figure}[H]
    \centering
    \includegraphics[width=\linewidth]{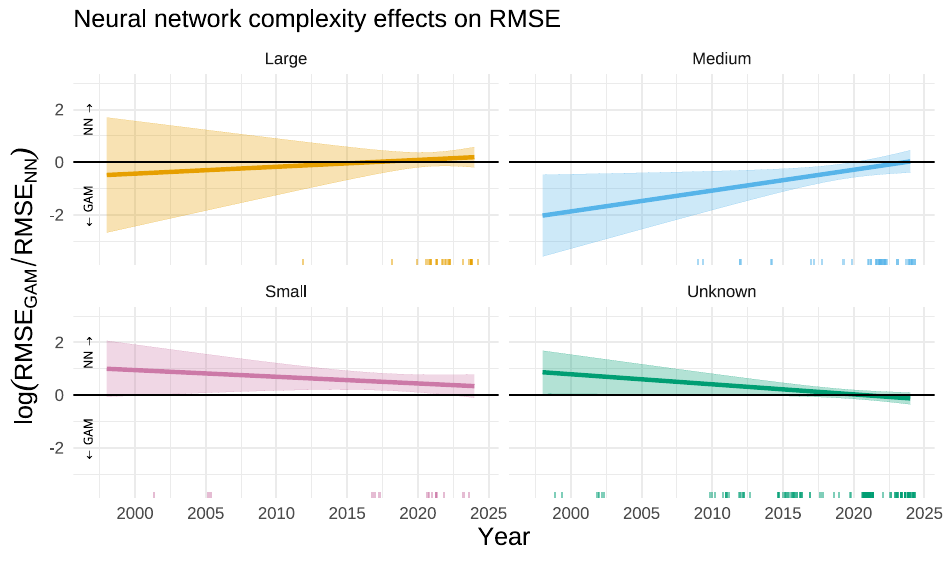}
    \caption{Marginal effects of the interaction between year and neural network complexity for the log RMSE ratio.
    Predicted values and 95\% confidence bands are derived from the mixed-effects model, averaging over the empirical distribution of other covariates.
    Rugs along the x-axis indicate the observations within each complexity category, highlighting where model estimates are supported by data. 
    Arrows indicate where neural networks are superior (above zero) and where GAMs are superior (below zero).}
    \label{fig: RMSE interaction plot}
\end{figure}

Figure~\ref{fig: RMSE interaction plot} displays the marginal effects of the interaction between year and neural network complexity for the log RMSE ratio produced using the \texttt{marginaleffects} R package \citep{mixedeffectsR}.
Predicted values and 95\% confidence bands were calculated using the empirical distribution of all other covariates, providing a marginalised view over the observed data.
This highlights how the performance of GAMs relative to neural networks changes over time across different complexity levels (where, again, positive values indicate that neural networks perform better).
The main finding is that the medium complexity neural networks are associated with worse performance relative to GAMs, but the effect is non-significant in the most recent five years. Indeed, for all complexity levels in 2024, the confidence bands include zero, indicating that neural network complexity is not associated with performance relative to GAMs at present (albeit other unreported neural network or GAM details could be).
Rug marks along the x-axis indicate the years for which there are observations for each complexity level, highlighting where data are available and model estimates are most reliable.

We also see from Table \ref{tab:full_model} that datasets of unknown sample size have a positive main effect  ($\hat{\beta}$ = 1.25, SE = 0.35) and a negative interaction with year ($\hat{\beta}$ = $-0.10$, SE = 0.03), both statistically significant.
Thus, at the 2011 baseline, neural networks appear to outperform GAMs in datasets with unknown sample size relative to the reference category (small), with this effect reducing over time. However, there is little practical implication here, as it is unclear how the datasets of unknown size differ from those where the sample size is reported.

The ICC is high (0.77), indicating substantial between-paper variability and supporting the appropriateness of a mixed-effects model in our analysis. It is possible that the large variability between papers could be explained by unreported features within papers that may influence model performance, e.g., data preprocessing and optimisation settings (see Section \ref{sec: unrecorded variables}).

\subsection{AUC metric analysis}\label{sec: AUC metric analysis}
The AUC analysis included 111 observations from 37 papers.
No outliers were identified, and all data points were included in the analysis.
For neural network complexity, the ``Large'' category was not represented in the data and hence omitted from the model. The ``Small'' and ``Medium'' categories were merged due to the minimal observations in each, and hence ``Small/Medium'' was used as the reference level.

Table~\ref{tab:full_model} presents the estimates from the AUC model, including all covariates and pairwise interactions with time.
The intercept is not statistically significant $\hat{\beta_0} = -1.81$ (SE = 1.61), so that, like with RMSE, there is no overall direction to the performance results, i.e., neural networks being better than GAMs, or vice versa.
However, we see that sample size and the number of predictors, along with their pairwise interactions with time, do have statistically significant effects, relative to the reference categories.

\begin{figure}[H]
    \centering
    \includegraphics[width=\linewidth]{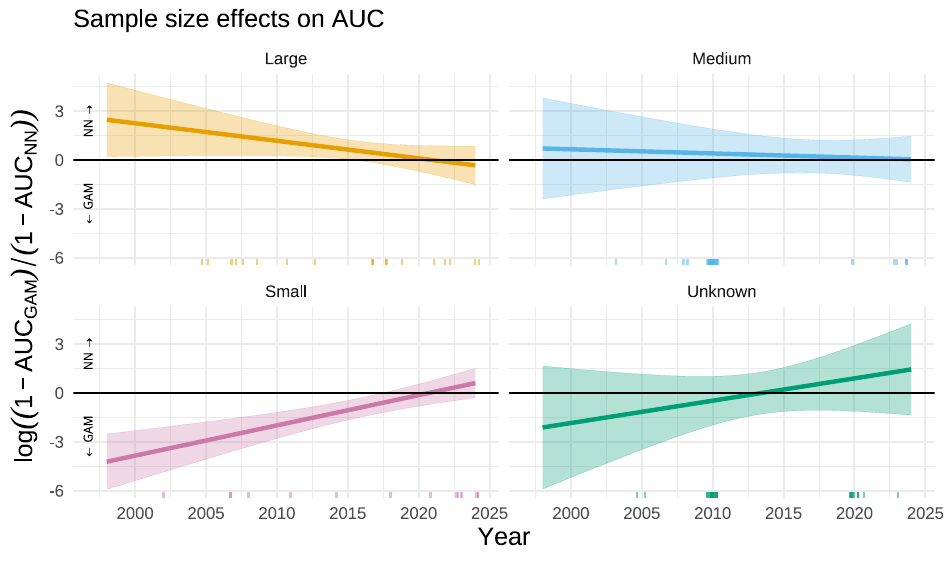}
    \caption{Marginal effects of the interaction between year and sample size for the log AUC loss ratio.
    Predicted values and 95\% confidence bands are derived from the mixed-effects model, averaging over the empirical distribution of other covariates.
    Rugs along the x-axis indicate the observations within each complexity category, highlighting where model estimates are supported by data.
    Arrows indicate where neural networks are superior (above zero) and where GAMs are superior (below zero).}
    \label{fig: AUC sample size interaction plot}
\end{figure}

The marginal effects plot for different sample sizes through time over the empirical distribution is seen in Figure~\ref{fig: AUC sample size interaction plot}. 
The confidence bands for medium and unknown sample sizes include zero throughout, indicating no association with performance differences between GAMs and neural networks.
In studies with small datasets prior to 2017, GAMs outperform neural networks, after which the difference is non-significant.
Conversely, in studies with large datasets, there is a slight tendency for neural networks to outperform GAMs, with the effect becoming non-significant by 2015. Thus, in recent years, there is no association between sample size and model performance. This could be due to advancements in optimisation/tuning procedures in both model classes over time, but the specifics of the procedures used are not available as previously discussed.

\begin{figure}[H]
    \centering
    \includegraphics[width=\linewidth]{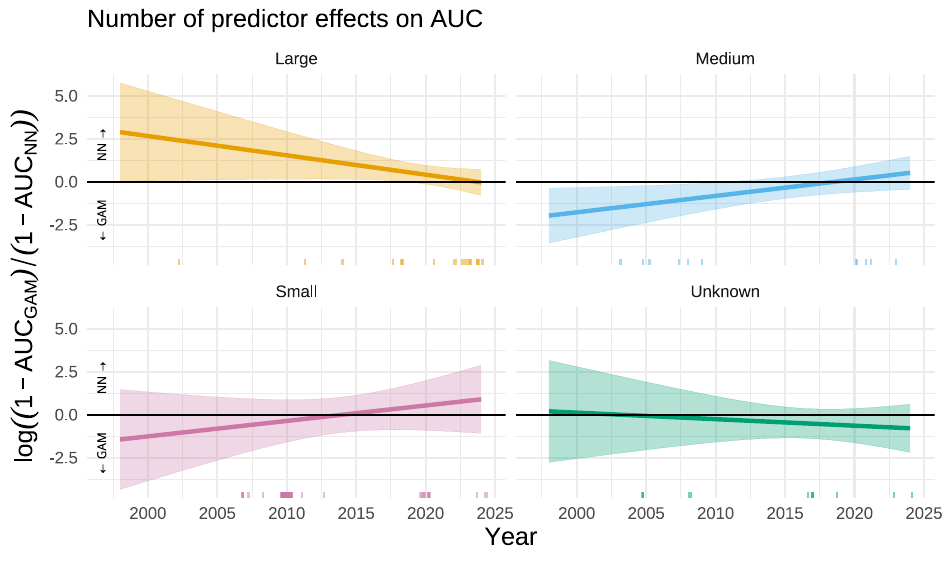}
    \caption{Marginal effects of the interaction between year and number of predictors for the log AUC loss ratio.
    Predicted values and 95\% confidence bands are derived from the mixed-effects model, averaging over the empirical distribution of other covariates.
    Rugs along the x-axis indicate the observations within each complexity category, highlighting where model estimates are supported by data.
    Arrows indicate where neural networks are superior (above zero) and where GAMs are superior (below zero).}
    \label{fig: AUC predictor interaction plot}
\end{figure}

The interaction between the number of predictors and year also has a statistically significant effect on the log AUC loss ratio.
The associated marginal effects are shown in Figure~\ref{fig: AUC predictor interaction plot}, where it is clear that the effect is relatively minimal.
Datasets with a medium number of predictors show the most notable change through time, slightly favouring GAMs over neural networks, but with no difference in recent years. 

Although our analysis provides some interesting signals that dataset dimensions (sample size and number of predictors) are associated with differences in model performance, the big variability in results across different papers is particularly noteworthy.
As with the RMSE analysis, the ICC is large here (ICC $= 0.97$).
Thus, results on datasets from the same paper are highly correlated with each other, which may indicate differences in the modelling approaches and datasets specific to individual papers that have not been quantified.

\subsection{$R^2$ metric analysis}\label{sec: R2 metric analysis}
Datasets utilising $R^2$ as a performance metric totalled 141, originating from 55 papers.
The log $R^2$ loss ratio was modelled in the same manner as the RMSE and AUC analyses, with the results provided in Table~\ref{tab:full_model}. In the case of the $R^2$ metric, there were no statistically significant effects, but the ICC was large ($0.78$) in line with the other two metrics.

\subsection{Sensitivity analysis}
\label{sec: sens analysis}
A series of sensitivity analyses were conducted to examine the stability of our performance-metric modelling under alternative data treatments and model specifications. We conducted a complete-case analysis, where cases with missing values in the sample size and number of predictors were excluded, and an analysis where these missing values were  imputed using multivariate imputation by chained equations via the \texttt{mice} package \citep{mice_package} in R. Note that, for both of these analyses, the neural network complexity variable was omitted due to its excessive amount of missingness. These analyses provide an alternative to the approach of including an ``Unknown'' category for missing values. We also carried out sub-analyses on the data from each of the four application domain areas separately, as an alternative to our previous overarching model wherein the domain appears as an additive effect. For these sub-analyses, we denote the fields as EAS (Environmental and Agricultural Sciences), EMCS (Engineering, Mathematics and Computer Science), HBS (Health and Biomedical Sciences), and OTH (Other). Across all of these variations, we find that the overall patterns in the results remain broadly consistent.

Table~\ref{tab:ICC_sens_analysis} displays the ICCs from all modelling approaches applied to all three metrics, where ``Main'' is the earlier model from Table~\ref{tab:full_model}. The ICC values remain high across the majority of scenarios, with values above 0.6 in almost all models. Interestingly, the lower ICC values relate to RMSE models, especially the complete data and HBS models. The generally large ICC values indicate significant differences between the results in different papers that cannot be explained by the available dataset-level variables. Moreover, the consistency in these high ICCs suggest that the clustering structure is an inherent feature of the literature (i.e., performance results are correlated within papers) rather than an artefact of any particular meta analysis model.

\begin{table}[p]
\centering
\caption{ICC}
\small
\label{tab:ICC_sens_analysis}
\begin{tabular}{lccccccc}
\toprule
 & \textbf{Main} & \textbf{Complete} & \textbf{Impute} & \textbf{EAS} & \textbf{EMCS} & \textbf{HBS} & \textbf{OTH}  \\
\midrule
\textbf{RMSE} & $0.76$ & $0.27$ & $0.99$ & $0.97$ & $0.64$ & $0.21$ &  $0.49$ \\
\textbf{AUC} & $0.97$ & $0.97$ & $0.74$ & $0.76$ & $0.88$ & $0.93$ &  $0.94$ \\
$\mathbf{R^2}$ & $0.78$ & $0.60$ & $0.77$ &  $0.64$ & $0.62$ & --- &  $0.95$  \\
\bottomrule
\end{tabular}
\end{table}

\begin{table}[p]
\centering
\caption{Likelihood ratio test p-values for each predictor across different meta analysis models for the three performance metrics.}
\footnotesize
\label{tab:complex-table}
\begin{threeparttable}
\begin{tabular}{llrrrrcccc }
\toprule
&  & \textbf{Main} & \textbf{Complete} & \textbf{Impute} & \textbf{EAS} & \textbf{EMCS} & \textbf{HBS} & \textbf{OTH} \\
\midrule

\multirow{3}{*}{\texttt{Year}} 
 & RMSE & 0.55 & 0.53 & 0.14 & 0.73 & 0.95 & 0.20 & 0.22 \\
 & AUC & 0.76 & 0.62 & 0.79 & 0.56 & 0.20 & 0.48 & 0.56\\
 & $R^2$ & 0.39 & 0.59 & 0.61 & 0.26 & 0.90 & --- & 0.97 \\

\midrule
\multirow{3}{*}{\texttt{Area}} 
 & RMSE & 0.39 & 0.91 & 0.87 & --- & --- & --- & ---  \\
 & AUC & 0.10 & 0.65 & 0.50 & --- & --- & --- & --- \\
 & $R^2$ & 0.71 & 0.89 & 0.90 & --- & --- & --- & --- \\

\midrule
\multirow{3}{*}{\texttt{Sample\_Size}} 
 & RMSE & $\mathbf{<0.01}$  & 0.50 & 0.75 & 0.27 & 0.40 & --- & 0.68 \\
 & AUC & $\mathbf{<0.01}$ & $\mathbf{<0.01}$ & 0.14 & \textbf{0.02} & --- & --- & 0.84\\
 & $R^2$ & 0.83 & 0.73 & 0.88 & 0.55 & 0.64 & --- & 0.81\\

\midrule
\multirow{3}{*}{\texttt{Predictors}} 
 & RMSE & 0.57 & 0.90 & 0.64 & 0.93 & 0.35 & --- & 0.95\\
 & AUC & \textbf{0.02} & 0.54 & \textbf{0.02} & 0.11 & --- & --- & ---\\
 & $R^2$ & 0.84 & 0.31 & 0.63 & 0.30 & 0.45 & --- & 0.95\\

\midrule
\multirow{3}{*}{\texttt{Complexity}} 
 & RMSE & $\mathbf{<0.01}$ & --- & --- & $\mathbf{<0.01}$ & 0.96 & --- & 0.07\\
 & AUC & 0.31 & --- & --- & 0.12 & --- & --- & ---\\
 & $R^2$ & 0.60 & --- & --- & 0.06 & 0.20 & --- & 0.48\\

\midrule
\multirow{3}{*}{\texttt{Year} $\times$ \texttt{Area}} 
 & RMSE & 0.69 & 0.92 & 0.38 & --- & --- & --- & ---\\
 & AUC & 0.73 & 0.17 & 0.43 & --- & --- & --- & --- \\
 & $R^2$ & 0.32 & 0.64 & 0.43 & --- & --- & --- & --- \\

\midrule
\multirow{3}{*}{\texttt{Year} $\times$ \texttt{Sample\_Size}} 
 & RMSE & \textbf{0.03} & 0.51 & 0.85 & 0.30 & 0.45 & --- & ---\\
 & AUC & $\mathbf{<0.01}$ & 0.10 & \textbf{0.03} & \textbf{0.03} & --- & --- & 0.52\\
 & $R^2$ & 0.92 & 0.70 & 0.86 & 0.51 & --- & --- & ---\\

\midrule
\multirow{3}{*}{\texttt{Year} $\times$ \texttt{Predictors}} 
 & RMSE & 0.69 & 0.71 & 0.66 & 0.83 & 0.31 & --- & ---\\
 & AUC & \textbf{0.01} & 0.17 & 0.06 & 0.13 & --- & --- & ---\\
 & $R^2$ & 0.81 & 0.63 & 0.70 & 0.12 & --- & --- & ---\\

\midrule
\multirow{3}{*}{\texttt{Year} $\times$ \texttt{Complexity}} 
 & RMSE & \textbf{0.03} & --- & --- & $\mathbf{<0.01}$ & 0.87 & --- & 0.09\\
 & AUC & 0.65 & --- & --- & 0.09 & --- & --- & ---\\
 & $R^2$ & 0.18 & --- & --- & 0.08 & --- & --- & ---\\

\bottomrule
\end{tabular}
\begin{tablenotes}
\item[]{P-values less than 0.05 are emboldened for clarity, and effects that were inestimable in any analysis approach are indicated using the ``---'' symbol.}
\end{tablenotes}
\end{threeparttable}
\end{table}
 
Significance of fixed effects was assessed using likelihood ratio tests for each of the seven analysis approaches, with resulting p-values provided in Table~\ref{tab:complex-table} (with the tables of regression coefficients deferred to \ref{app: sensitivity}). The sample size effect over time is most often significant in the AUC models, whereas the effect of the number of predictors over time is only significant in the Main AUC model (which was a weaker effect from Figure~\ref{fig: AUC predictor interaction plot}). The imputation-based analysis identifies the effect of the sample size over time and the main effect of predictors (but where the time interaction has a p-value of $0.06$) in the AUC model, while the complete-case analysis only identifies the main effect of sample size. The key finding from the previous Main RMSE model was that the neural network complexity effect was significant. As this variable was removed from the complete-case and imputation-based analyses, it cannot appear in those models. However, its effect is significant in the RMSE model for the EAS papers. The AUC model for the EAS papers also identifies the effect of sample size as per the other AUC models, but not the effect of predictors (which, again, was the weaker effect). None of the other fields identify significant effects, but recall from Table~\ref{tab: paper areas} that the number of papers in each of these is smaller than for EAS (see also Table \ref{tab:paper_data_sensitity} in \ref{app: sensitivity}); thus, the power to detect effects will be lower in each of these fields when considered individually. As for the $R^2$ metric, no variables are statistically significant in any analysis, in line with the Main $R^2$ model from before. We also conducted a version of the complete case analysis with results weighted by the sample size, which did not yield further insights (see \ref{app: sensitivity}). Overall, our sensitivity analysis demonstrates that the main findings are relatively robust across variations in data handling and model specification.

\section{Conclusion}\label{sec: Conclusion}
This is the first systematic review of papers that compare GAMs and neural networks on real tabular datasets. In total, our review includes results from 430 datasets across 143 different papers, where we have also conducted an analysis of the performance metrics within these papers. Our work offers new insights into how these two modelling frameworks have been applied, assessed, and reported on within the literature.

Summarising the papers by year of publication, journal title and ranking, application area, problem type (regression/classification), sample size, number of predictors, and neural network type and complexity 
provides answers to RQ1 on the characteristics of papers comparing GAMs and neural networks. We found that the majority of papers in this area were published in top-ranking journals (approx.~70\% in quartile one). No individual journal appeared more than three times, indicating a diversity of sub-disciplines represented from four broad domain areas (Environmental \& Agricultural Sciences; Engineering, Mathematics \& Computer Science; Health \& Biomedical Sciences; and Other.) There has also been a recent surge of papers that compare GAMs and neural networks, particularly in the last five years. These findings highlight the growing interest in comparing these models across the scientific community, and timeliness of the current review. 

Our study was not specific to any particular neural network architecture based on our search string or inclusion criteria (see Section~\ref{sec: Methods}). However, the abundance of studies using MLPs (approx.~90\%) was particularly noteworthy, and 70\% of these were shallow architectures consisting of a single hidden layer.
While alternative neural network structures have been shown to improve on MLPs on tabular datasets, e.g., TabNet and Transformers \citep{arik2021tabnet}, comparisons with GAMs are lacking, and would be an interesting area of future research. Other recent studies show that tree-based procedures (e.g., random forests and XGBoost) outperform neural networks on tabular datasets \citep{borisov2022deep, shwartz2022tabular}. On the other hand, \cite{kadra2021well} indicate that well-regularised MLPs can outperform other neural network structures and tree-based methods, suggesting that MLPs are a competitive baseline if appropriately regularised. Thus, MLPs remain relevant and are clearly very frequently used across diverse areas of applied research, but further comparison studies including modern machine learning methods, MLPs, and GAMs are of interest.

A very notable finding from RQ1 on the characteristics of papers carrying out these comparisons is substantial heterogeneity in how datasets and models are reported. 
Important details such as sample size, number of predictors, and neural network complexity were frequently unreported, with missingness of approximately 20\% for the former two variables, and 65\% for the latter. Other details on data preprocessing steps, optimisation algorithms, and hyperparameter tuning were inconsistently reported to the extent that they could not be captured in this review. It should be highlighted that missing information is a common feature of systematic reviews due to differences that will naturally arise in how different authors report results \citep{andaur2022completeness}. However, the lack of standardised reporting within machine learning research has previously been identified due to its impact on transparency and reproducibility \citep{mitchell2019model}. These reporting issues have been confirmed within our systematic review across many diverse research articles; the unreported information may also partly explain the high intra-paper correlation observed in our mixed-effects models. A particularly interesting future direction would be gathering datasets identified within this review, and independently analysing them according to a standardised modelling protocol, e.g., with
consistent preprocessing, optimisation, tuning, and evaluation procedures to remove the effects of the different approaches taken in different papers.

We also extracted and analysed performance outcomes across the most commonly reported metrics (RMSE, $R^2$, and AUC) to investigate both RQ2 (on whether or not one model outperforms the other) and RQ3 (and whether or not performance differences can be explained using reported variables).
On average, there was no evidence of a statistically significant difference between the performance of GAMs and neural networks across datasets for all three metrics based on the values of the intercepts of our meta analysis models (RQ2). That is to say, there is no evidence of consistent superiority of one model class over the other.
This aligns with the claims of other authors, for example, \cite{rudin2019stop} says that no single algorithm consistently outperforms others on tabular data, with differences often being small compared to tuning effects.
Indeed, Rudin suggests that interpretable models can perform similarly to complex ones, though publication bias often exaggerates performance gaps.

Considering the effects of paper and dataset characteristics revealed some potential associations with differences in the performance of GAMs and neural networks (RQ3). For RMSE, model complexity and its interaction with year was statistically significant. GAMs performed better than neural networks in studies where the neural network was medium-sized (5 to 10 parameters), but this effect has diminished in recent years and is no longer statistically significant. For AUC, the effects of dataset dimensions (sample size and the number of predictors) showed that, while neural networks are associated with improved performance in larger higher-dimensional datasets, the performance gap has narrowed over time. In recent years, there were no statistically significant effects associated with performance differences between GAMs and neural networks. While being conscious of the unreported information as discussed above, these results indicate that GAMs and neural networks are currently well matched across many different analyses conducted in the literature.

Our findings offer important insights for future research.
First, GAMs remain a robust and interpretable choice in many applied settings and should not be dismissed simply due to the perceived superiority of more complex models. Indeed, GAMs are well suited to scientific research, where explainability and transparency are paramount; note that explainable machine learning is now a large field \citep{montavon2019explainable, roscher2020explainable, burkart2021survey}.
Second, in the early-to-mid-2000s, while the performance advantage of models was associated with neural network complexity and dataset dimension, these factors no longer explain performance differences. Thus, it would appear that modern GAMs and neural networks cannot be distinguished based on these features, but it is possible that other features of optimisation and hyperparameter tuning may separate the two. Thus, future research should focus on adjusting for these additional modelling choices when conducting comparisons rather than only using one specific implementation of each model. Third, the strong intra-paper correlation (high ICC values across all metrics) demonstrates the necessity of accounting for paper-level variation when comparing models.
Given the large variability in model performance across different papers, and inconsistency in reporting, we  suggest that more standardised reporting of results should become the norm. An example of such a reporting standard is the ``model card'' proposed in \cite{mitchell2019model}. This is particularly important as such models are increasingly being deployed (and automated) in mainstream applications, where transparency is becoming a focus (e.g., the EU's AI Act).

In conclusion, our study provides a comprehensive assessment of the performance of GAMs versus neural networks across a diverse set of applications using tabular data. We have also made the data collected available at \url{https://github.com/jessicadoohan/gam-vs-nn-review}, so that others can expand on this study in the future.
While neural networks have shown impressive performance on complex (non-tabular) datasets, GAMs continue to offer competitive performance with added interpretability and transparency on tabular datasets.
Neural networks and GAMs should be considered complementary tools within the modern data-science toolkit.
Choosing between them requires careful consideration of the value placed on interpretability versus predictive gains.
Notably, recent research has been conducted on combining GAMs and neural networks within hybrid models \citep{Yang2021, NAM_timeseries2023, deepregression_R_paper_2023, rugamer2024semi}.

\section*{Acknowledgement}
This publication has emanated from research conducted with the financial support of Taighde Éireann – Research Ireland under Grant number 18/CRT/6049. For the purpose of Open Access, the author has applied a CC BY public copyright licence to any Author Accepted Manuscript version arising from this submission

\bibliographystyle{apalike}
\bibliography{main}

\newpage
\appendix


\section{Alternative model formulation}\label{app: model alternative}
In Section~\ref{sec: RMSE analysis}, the log RMSE ratio (GAM/NN) is modelled as
\begin{equation}
\log(\mathrm{RMSE}_{\mathrm{ratio}}) \sim \beta_0 + \texttt{Year} + \texttt{Area} + \dots + (\texttt{Year} \times \texttt{Complexity}) + ( 1 | \texttt{Paper\_ID} ), 
\end{equation}
which is equivalent to modelling
\begin{multline}
\log(\mathrm{RMSE}_{\mathrm{GAM}}) \sim \beta_0 +  \beta_1 \log(\mathrm{RMSE}_{\mathrm{NN}}) + \texttt{Year} + \texttt{Area} + \\
\dots + (\texttt{Year} \times \texttt{Complexity}) + ( 1 | \texttt{Paper\_ID} ), 
\end{multline}
where $\beta_1$ = 1.
More generally, one may estimate $\beta_1$ from the data.
In this more general modelling framework, we model $\log(\mathrm{RMSE_{GAM}})$ and include $\log(\mathrm{RMSE_{NN}})$  as a covariate along with publication year, domain area, sample size, number of predictors, and neural network complexity.
Results from this model are reported in Table~\ref{tab:full_model_alternative}. 
(Models for AUC and $R^2$ are also shown.)
The estimated coefficient for the covariate $\log(\mathrm{RMSE_{NN}})$ is $\hat{\beta_1} = 0.99$ (SE = 0.01).
There is no evidence that this estimate is statistically significantly different from 1 at the 5\% level.
This implies that the RMSE of GAMs and neural networks changes in a one-to-one proportion. 
Importantly, the set of significant predictors and their interpretations remain identical to those in Table~\ref{tab:full_model}.
\begin{table}[H]
\centering
\caption{Estimates from alternative mixed-effects model.} 
\label{tab:full_model_alternative}
\begin{threeparttable}
{\footnotesize
\begin{tabular}{lccc}
  \toprule
   & \textbf{RMSE} & \textbf{AUC} & $\boldsymbol{R}^{\boldsymbol{2}}$\\ 
  \textbf{Term} & \textbf{Estimate (SE)} &  \textbf{Estimate (SE)}  &  \textbf{Estimate (SE)} \\ 
  \midrule
(Intercept) & $\phantom{-}$0.11 (0.27)\phantom{$^*$} & $-$1.84 (1.64)\phantom{$^*$} & $-$0.71 (0.97)\phantom{$^*$} \\ 

\texttt{$\log(\mathrm{Metric_{NN}})$} & $\phantom{-}$0.99 (0.01)$^*$ & $\phantom{-}$1.03 (0.04)$^*$ & $\phantom{-}$0.60 (0.06)$^*$ \\ 

\texttt{Year} & $\phantom{-}$0.02 (0.03)\phantom{$^*$} & $\phantom{-}$0.36 (0.21)\phantom{$^*$} & $-$0.02 (0.09)\phantom{$^*$} \\ 
 
\texttt{Area$_\texttt{Env/Agri}$} & $\phantom{-}$0.26 (0.30)\phantom{$^*$} & $-$0.72 (1.56)\phantom{$^*$} & $\phantom{-}$0.33 (0.87)\phantom{$^*$} \\
\texttt{Area$_\texttt{Health/Bio}$} & $-$0.35 (0.45)\phantom{$^*$} & $\phantom{-}$1.34 (1.30)\phantom{$^*$} & $\phantom{-}$0.76 (0.92)\phantom{$^*$} \\
\texttt{Area$_\texttt{Other}$} & $-$0.12 (0.41)\phantom{$^*$} & $-$0.40 (1.37)\phantom{$^*$} & $\phantom{-}$0.13 (1.11)\phantom{$^*$} \\

\texttt{Sample\_Size$_{medium}$} & $\phantom{-}$0.33 (0.33)\phantom{$^*$} & $\phantom{-}$3.63 (0.27)$^*$ & $\phantom{-}$0.50 (0.52)\phantom{$^*$} \\ 
\texttt{Sample\_Size$_{large}$} & $\phantom{-}$0.80 (0.44)\phantom{$^*$} & $\phantom{-}$2.81 (0.25)$^*$ & $\phantom{-}$0.69 (1.16)\phantom{$^*$} \\ 
\texttt{Sample\_Size$_{unknown}$} & $\phantom{-}$1.28 (0.35)$^*$ & $\phantom{-}$2.61 (0.78)$^*$ & $\phantom{-}$0.44 (0.75)\phantom{$^*$} \\ 
 
\texttt{Predictors$_{medium}$} & $\phantom{-}$0.26 (0.22)\phantom{$^*$} & $-$0.08 (0.24)\phantom{$^*$} & $\phantom{-}$0.09 (0.37)\phantom{$^*$} \\ 
\texttt{Predictors$_{large}$} & $\phantom{-}$0.37 (0.34)\phantom{$^*$} & $\phantom{-}$2.25 (0.72)$^*$ & $\phantom{-}$0.08 (0.72)\phantom{$^*$} \\ 
\texttt{Predictors$_{unknown}$} & $\phantom{-}$0.10 (0.38)\phantom{$^*$} & $\phantom{-}$0.88 (1.06)\phantom{$^*$} & $-$0.20 (0.68)\phantom{$^*$} \\ 
 
\texttt{Complexity$_{medium}$} & $-$1.74 (0.38)$^*$ &  & $-$0.51 (0.54)\phantom{$^*$} \\ 
\texttt{Complexity$_{large}$} & $-$1.46 (0.51)$^*$ &  & $-$0.29 (0.78)\phantom{$^*$} \\ 
\texttt{Complexity$_{unknown}$} & $-$0.54 (0.29)\phantom{$^*$} & $-$1.05 (1.04)\phantom{$^*$} & $\phantom{-}$0.08 (0.46)\phantom{$^*$} \\

\texttt{Year} × \texttt{Area$_\texttt{Env/Agri}$} & $-$0.02 (0.03)\phantom{$^*$} & $\phantom{-}$0.02 (0.17)\phantom{$^*$} & $\phantom{-}$0.00 (0.08)\phantom{$^*$} \\
\texttt{Year} × \texttt{Area$_\texttt{Health/Bio}$} & $\phantom{-}$0.01 (0.05)\phantom{$^*$} & $-$0.04 (0.15)\phantom{$^*$} &  \\ 
\texttt{Year} × \texttt{Area$_\texttt{Other}$} & $\phantom{-}$0.02 (0.04)\phantom{$^*$} & $\phantom{-}$0.05 (0.15)\phantom{$^*$} & $\phantom{-}$0.04 (0.11)\phantom{$^*$} \\ 

\texttt{Year} × \texttt{Sample\_Size$_{medium}$} & $-$0.02 (0.03)\phantom{$^*$} & $-$0.32 (0.02)$^*$ & $-$0.01 (0.05)\phantom{$^*$} \\ 
\texttt{Year} × \texttt{Sample\_Size$_{large}$} & $-$0.06 (0.04)\phantom{$^*$} & $-$0.30 (0.05)$^*$ & $-$0.04 (0.10)\phantom{$^*$} \\
\texttt{Year} × \texttt{Sample\_Size$_{unknown}$} & $-$0.10 (0.03)$^*$ & $-$0.21 (0.09)$^*$ & $-$0.01 (0.07)\phantom{$^*$} \\ 

\texttt{Year} × \texttt{Predictors$_{medium}$} & $-$0.03 (0.02)\phantom{$^*$} & $-$0.03 (0.06)\phantom{$^*$} & $-$0.03 (0.05)\phantom{$^*$} \\ 
\texttt{Year} × \texttt{Predictors$_{large}$} & $-$0.04 (0.04)\phantom{$^*$} & $-$0.25 (0.09)$^*$ & $\phantom{-}$0.00 (0.07)\phantom{$^*$} \\ 
\texttt{Year} × \texttt{Predictors$_{unknown}$} & $-$0.01 (0.04)\phantom{$^*$} & $-$0.16 (0.13)\phantom{$^*$} & $\phantom{-}$0.06 (0.07)\phantom{$^*$} \\

\texttt{Year} × \texttt{Complexity$_{medium}$} & $\phantom{-}$0.11 (0.04)$^*$ &  & $\phantom{-}$0.05 (0.06)\phantom{$^*$} \\ 
\texttt{Year} × \texttt{Complexity$_{large}$} & $\phantom{-}$0.10 (0.05)$^*$ &  & $\phantom{-}$0.01 (0.07)\phantom{$^*$} \\ 
\texttt{Year} × \texttt{Complexity$_{unknown}$} & $\phantom{-}$0.01 (0.03)\phantom{$^*$} & $-$0.06 (0.13)\phantom{$^*$} & $-$0.03 (0.05)\phantom{$^*$} \\

$\sigma^2_{paper}$ & $\phantom{-}$0.26 \phantom{(0.00)$^*$} & $\phantom{-}$1.10 \phantom{(0.00)$^*$} & $\phantom{-}$0.62 \phantom{(0.00)$^*$}\\ 
$\sigma^2_{residual}$ & $\phantom{-}$0.08 \phantom{(0.00)$^*$} & $\phantom{-}$0.03 \phantom{(0.00)$^*$} & $\phantom{-}$0.16 \phantom{(0.00)$^*$}\\ 
ICC & $\phantom{-}$0.76 \phantom{(0.00)$^*$} & $\phantom{-}$0.97 \phantom{(0.00)$^*$} & $\phantom{-}$0.79 \phantom{(0.00)$^*$} \\
\bottomrule
\end{tabular}}
{\footnotesize\begin{tablenotes}
\item[]{($^{*}$) indicates statistical significance at the 5\% level.}
\end{tablenotes}}
\end{threeparttable}
\end{table}

Repeating this alternative modelling structure for AUC, we come to similar conclusions with the estimated coefficient for $\log(\mathrm{(1-AUC_{NN})})$ being $\hat{\beta_1} = 1.03$ (SE = 0.04), which is not statistically significantly different from 1 at the 5\% level.
The set of significant predictors is the same as the modelling framework in the main text.
For $R^2$, the estimate for $\log(\mathrm{(1-R^2_{NN})})$ is $\hat{\beta_1} = 0.60$ (SE = 0.06), which is statistically significantly different from 1 at the 5\% level.
However, similar to the model in the main text, there is no evidence of any predictors having an effect on the response.

\section{Sensitivity analysis} \label{app: sensitivity}
This appendix presents additional sensitivity analyses to assess the robustness of the main findings across alternative model specifications, subsets of the data, and weighting by sample size.
Table \ref{tab:paper_data_sensitity} summarises how papers and datasets are distributed across model configurations for each performance metric.

\begin{table}[H]
\centering
\caption{Number of papers and datasets for each model configuration.}

\label{tab:paper_data_sensitity}
\begin{threeparttable}
\begin{tabular}{lr@{\,}rr@{\,}rrr@{\,}rr@{\,}rr@{\,}rr@{\,}rr@{\,}r}
\toprule
 & \multicolumn{2}{c}{\textbf{Main}} & \multicolumn{2}{c}{\textbf{Complete\!\!\!\!\!\!}} && \multicolumn{2}{c}{\textbf{Impute}} & \multicolumn{2}{c}{\textbf{EAS}} & \multicolumn{2}{c}{\textbf{EMCS}} & \multicolumn{2}{c}{\textbf{HBS}} & \multicolumn{2}{c}{\textbf{OTH}}  \\
\midrule
\textbf{RMSE}  &  78 & (227) & 57 & (149) && 78 & (227)  & 34 & (66) & 26 & (112) &  4 & (14)  & 14 & (35)\\
\textbf{AUC}   & 37  & (111) & 27 & (65)  && 37 & (111)  & 17 & (81) & 5  &   (6) &  9 & (10)  &  6 & (14) \\
$\mathbf{R^2}$ & 55  & (141) & 40 & (98)  && 55 & (141)  & 30 & (60) & 11 & (44)  &  1 &  (6)  & 12 & (30) \\
\bottomrule
\end{tabular}
\begin{tablenotes}
\footnotesize
\item[] Papers are shown first with datasets shown in parentheses.  
\end{tablenotes}
\end{threeparttable}
\end{table}

Table \ref{tab:sens_complete_model} reports mixed-effects model estimates based on the complete-case dataset, where observations with unknown sample sizes and number of predictors are removed. Table \ref{tab:sense_impute_model} shows estimates based on imputed data, i.e., where the unknown sample sizes and number of predictors are imputed. In both of these modelling approaches, the neural network complexity variable was removed due to its very high level of missingness.

Tables \ref{tab:env_models}, \ref{tab:eng_models}, \ref{tab:health_models}, and \ref{tab:other_models} respectively show the estimated coefficients from models fitted to each of the four domain areas separately, i.e., Environmental and Agricultural Science (EAS), Engineering, Mathematics and Computer Science (EMCS),  Health and Biomedical Science (HBS), and Other (OTH).

Lastly, Table~\ref{tab:sens_weighted} fits a model where the observations are weighted by the sample size rather than including sample size as a covariate. This gives results from applications to larger datasets in the literature a larger weight in the meta analysis model. Note that this is conducted within the complete case analysis, where sample size is a known numeric variable, and, hence, the complexity variable was not included.

In most cases, individual coefficients are not statistically significant within these models. This does not necessarily mean that the overall covariates effects are not important, but rather that the differences compared to the reference categories are not statistically significant; overall covariate p-values were provided in Table \ref{tab:complex-table} in the main paper. Note also that the number of papers analysed is naturally smaller in the complete case analysis and the various subset analyses, such that power to detect effects will be reduced. Interestingly, the EAS subset analysis (the largest subset) is closest to the full analysis in the main paper, i.e., the complexity-by-year interaction is statistically significant in RMSE model, and the sample size effect is statistically significant in the AUC model (but not its interaction with year, or the effect of predictor variables). In the analysis weighted by sample size, only one coefficient within the predictors-by-year interaction in the AUC model is statistically significant.

\begin{table}[p]
\centering
\caption{Estimates from mixed-effects models using the complete dataset.} 
\label{tab:sens_complete_model}
\begin{threeparttable}
{\small
\begin{tabular}{lccc}
   \toprule
   & \textbf{RMSE} & \textbf{AUC} & $\boldsymbol{R}^{\boldsymbol{2}}$\\ 
  \textbf{Term} & \textbf{Estimate (SE)} &  \textbf{Estimate (SE)} & \textbf{Estimate (SE)} \\ 
  \midrule
(Intercept) & $-0.02$ (0.19)\phantom{$^*$} & $\phantom{-}0.85$ (0.85)\phantom{$^*$} & $-0.68$ (0.86)\phantom{$^*$} \\ 

\texttt{Year} & $\phantom{-}0.00$ (0.02)\phantom{$^*$} & $\phantom{-}0.05$ (0.10)\phantom{$^*$} & $\phantom{-}0.04$ (0.08)\phantom{$^*$} \\ 
 
\texttt{Area$_\texttt{Env/Agri}$} & $-0.10$ (0.22)\phantom{$^*$} & $-0.57$ (0.72)\phantom{$^*$} & $\phantom{-}0.57$ (0.77)\phantom{$^*$} \\
\texttt{Area$_\texttt{Health/Bio}$} & $\phantom{-}0.07$ (0.45)\phantom{$^*$} & $-0.17$ (0.73)\phantom{$^*$} & $-0.13$ (0.82)\phantom{$^*$} \\
\texttt{Area$_\texttt{Other}$} & $-0.19$ (0.41)\phantom{$^*$} & $-0.48$ (0.75)\phantom{$^*$} & $\phantom{-}0.33$ (1.10)\phantom{$^*$} \\

\texttt{Sample\_Size$_{medium}$} & $\phantom{-}0.30$ (0.26)\phantom{$^*$} & $-0.10$ (0.53)\phantom{$^*$} & $-0.17$ (0.47)\phantom{$^*$} \\ 
\texttt{Sample\_Size$_{large}$} & $\phantom{-}0.21$ (0.38)\phantom{$^*$} & $-0.83$ (0.52)\phantom{$^*$} & $-1.65$ (2.51)\phantom{$^*$} \\ 
 
\texttt{Predictors$_{medium}$} & $\phantom{-}0.05$ (0.20)\phantom{$^*$} & $-0.17$ (0.16)\phantom{$^*$} & $\phantom{-}0.65$ (0.46)\phantom{$^*$} \\ 
\texttt{Predictors$_{large}$} & $-0.12$ (0.38)\phantom{$^*$} & $\phantom{-}0.04$ (0.50)\phantom{$^*$} & $\phantom{-}1.99$ (2.51)\phantom{$^*$} \\ 
 
\texttt{Year} × \texttt{Area$_\texttt{Env/Agri}$} & $\phantom{-}0.00$ (0.02)\phantom{$^*$} & $-0.02$ (0.08)\phantom{$^*$} & $-0.05$ (0.07)\phantom{$^*$} \\
\texttt{Year} × \texttt{Area$_\texttt{Health/Bio}$} & $-0.01$ (0.05)\phantom{$^*$} & $\phantom{-}0.09$ (0.08)\phantom{$^*$} & --- \\ 
\texttt{Year} × \texttt{Area$_\texttt{Other}$} & $\phantom{-}0.03$ (0.04)\phantom{$^*$} & $\phantom{-}0.02$ (0.08)\phantom{$^*$} & $\phantom{-}0.01$ (0.11)\phantom{$^*$} \\ 
 
\texttt{Year} × \texttt{Sample\_Size$_{medium}$} & $-0.03$ (0.03)\phantom{$^*$} & $-0.03$ (0.04)\phantom{$^*$} & $\phantom{-}0.03$ (0.05)\phantom{$^*$} \\ 
\texttt{Year} × \texttt{Sample\_Size$_{large}$} & $-0.01$ (0.04)\phantom{$^*$} & $\phantom{-}0.02$ (0.05)\phantom{$^*$} & $\phantom{-}0.16$ (0.22)\phantom{$^*$} \\ 

\texttt{Year} × \texttt{Predictors$_{medium}$} & $-0.01$ (0.02)\phantom{$^*$} & $-0.06$ (0.04)\phantom{$^*$} & $-0.03$ (0.05)\phantom{$^*$} \\ 
\texttt{Year} × \texttt{Predictors$_{large}$} & $\phantom{-}0.01$ (0.04)\phantom{$^*$} & $-0.11$ (0.06)\phantom{$^*$} & $-0.17$ (0.22)\phantom{$^*$} \\

   \bottomrule
\end{tabular}}
 {\footnotesize\begin{tablenotes}
	\item[]{($^{*}$) indicates statistical significance at the 5\% level.}
	\end{tablenotes}}
\end{threeparttable}
\end{table}

\begin{table}[p]
\centering
\caption{Estimates from mixed-effects models pooled from imputed datasets.} 
\label{tab:sense_impute_model}
\begin{threeparttable}
{\small
\begin{tabular}{lccc}
   \toprule
   & \textbf{RMSE} & \textbf{AUC} & $\boldsymbol{R}^{\boldsymbol{2}}$\\ 
  \textbf{Term} & \textbf{Estimate (SE)} &  \textbf{Estimate (SE)} & \textbf{Estimate (SE)} \\ 
  \midrule
(Intercept) & $-0.22$ (1.19)\phantom{$^*$} & $-0.10$ (0.93)\phantom{$^*$} & $-0.38$ (0.95)\phantom{$^*$} \\ 

\texttt{Year} & $\phantom{-}0.11$ (0.12)\phantom{$^*$} & $\phantom{-}0.04$ (0.13)\phantom{$^*$} & $\phantom{-}0.05$ (0.09)\phantom{$^*$} \\ 
 
\texttt{Area$_\texttt{Env/Agri}$} & $\phantom{-}0.32$ (1.41)\phantom{$^*$} & $-0.33$ (0.91)\phantom{$^*$} & $\phantom{-}0.42$ (0.90)\phantom{$^*$} \\
\texttt{Area$_\texttt{Health/Bio}$} & $\phantom{-}1.49$ (1.99)\phantom{$^*$} & $\phantom{-}0.12$ (0.94)\phantom{$^*$} & $-0.42$ (0.96)\phantom{$^*$} \\
\texttt{Area$_\texttt{Other}$} & $-0.05$ (1.84)\phantom{$^*$} & $-0.32$ (0.99)\phantom{$^*$} & $\phantom{-}0.03$ (1.17)\phantom{$^*$} \\

\texttt{Sample\_Size$_{medium}$} & $\phantom{-}0.03$ (0.38)\phantom{$^*$} & $\phantom{-}0.30$ (0.27)\phantom{$^*$} & $\phantom{-}0.07$ (0.38)\phantom{$^*$} \\ 
\texttt{Sample\_Size$_{large}$} & $\phantom{-}0.40$ (0.66)\phantom{$^*$} & $\phantom{-}0.21$ (0.26)\phantom{$^*$} & $\phantom{-}0.14$ (0.82)\phantom{$^*$} \\ 
 
\texttt{Predictors$_{medium}$} & $-0.01$ (0.32)\phantom{$^*$} & $-0.06$ (0.39)\phantom{$^*$} & $\phantom{-}0.31$ (0.39)\phantom{$^*$} \\ 
\texttt{Predictors$_{large}$} & $\phantom{-}0.15$ (0.56)\phantom{$^*$} & $\phantom{-}0.39$ (1.14)\phantom{$^*$} & $\phantom{-}0.35$ (0.69)\phantom{$^*$} \\ 
 
\texttt{Year} × \texttt{Area$_\texttt{Env/Agri}$} & $-0.13$ (0.14)\phantom{$^*$} & $-0.01$ (0.11)\phantom{$^*$} & $-0.07$ (0.08)\phantom{$^*$} \\
\texttt{Year} × \texttt{Area$_\texttt{Health/Bio}$} & $\phantom{-}0.20$ (0.21)\phantom{$^*$} & $\phantom{-}0.05$ (0.11)\phantom{$^*$} &  ---\\ 
\texttt{Year} × \texttt{Area$_\texttt{Other}$} & $-0.08$ (0.18)\phantom{$^*$} & $\phantom{-}0.02$ (0.11)\phantom{$^*$} & $\phantom{-}0.02$ (0.11)\phantom{$^*$} \\ 
 
\texttt{Year} × \texttt{Sample\_Size$_{medium}$} & $\phantom{-}0.01$ (0.04)\phantom{$^*$} & $-0.05$ (0.03)\phantom{$^*$} & $\phantom{-}0.01$ (0.04)\phantom{$^*$} \\ 
\texttt{Year} × \texttt{Sample\_Size$_{large}$} & $-0.02$ (0.06)\phantom{$^*$} & $-0.04$ (0.04)\phantom{$^*$} & $\phantom{-}0.00$ (0.07)\phantom{$^*$} \\ 

\texttt{Year} × \texttt{Predictors$_{medium}$} & $\phantom{-}0.00$ (0.03)\phantom{$^*$} & $-0.02$ (0.06)\phantom{$^*$} & $-0.02$ (0.04)\phantom{$^*$} \\ 
\texttt{Year} × \texttt{Predictors$_{large}$} & $-0.02$ (0.05)\phantom{$^*$} & $-0.06$ (0.14)\phantom{$^*$} & $-0.04$ (0.07)\phantom{$^*$} \\
   \bottomrule
\end{tabular}}
 {\footnotesize\begin{tablenotes}
	\item[]{($^{*}$) indicates statistical significance at the 5\% level.}
	\end{tablenotes}}
\end{threeparttable}
\end{table}

\begin{table}[p]
\centering
\caption{Estimates from mixed-effects models for the EAS subset analysis.} 
\label{tab:env_models}
\begin{threeparttable}
{\small
\begin{tabular}{lccc}
   \toprule
   & \textbf{RMSE} & \textbf{AUC} & $\boldsymbol{R}^{\boldsymbol{2}}$\\ 
  \textbf{Term} & \textbf{Estimate (SE)} &  \textbf{Estimate (SE)} & \textbf{Estimate (SE)} \\ 
  \midrule
(Intercept) & $\phantom{-}1.69$ (0.64)$^*$ & $-2.33$ (0.53)\phantom{$^*$} & $\phantom{-}0.14$ (0.40)\phantom{$^*$} \\ 

\texttt{Year} & $\phantom{-}0.08$ (0.09)\phantom{$^*$} & $\phantom{-}0.47$ (0.15)\phantom{$^*$} & $\phantom{-}0.02$ (0.05)\phantom{$^*$} \\ 
 
\texttt{Sample\_Size$_{medium}$} & $\phantom{-}2.88$ (1.83)\phantom{$^*$} & $\phantom{-}5.59$ (1.79)\phantom{$^*$} & $\phantom{-}0.81$ (0.56)\phantom{$^*$} \\
\texttt{Sample\_Size$_{large}$} & $-0.84$ (2.88)\phantom{$^*$} & $-0.08$ (2.71)\phantom{$^*$} & $\phantom{-}0.31$ (1.33)\phantom{$^*$} \\
\texttt{Sample\_Size$_{unknown}$} & $\phantom{-}1.86$ (1.36)\phantom{$^*$} & $\phantom{-}2.68$ (0.40)$^*$ & $-0.06$ (0.61)\phantom{$^*$} \\

\texttt{Predictors$_{medium}$} & $-0.02$ (0.28)\phantom{$^*$} & $-15.39$ (4.80)\phantom{$^*$} & $\phantom{-}0.16$ (0.33)\phantom{$^*$} \\ 
\texttt{Predictors$_{large}$} & $\phantom{-}1.04$ (1.61)\phantom{$^*$} & $-12.04$ (5.90)\phantom{$^*$} & $\phantom{-}1.30$ (0.80)\phantom{$^*$} \\ 
\texttt{Predictors$_{unknown}$} & $\phantom{-}0.04$ (3.10)\phantom{$^*$} & $\phantom{-}2.83$ (1.72)\phantom{$^*$} & $-0.24$ (1.07)\phantom{$^*$} \\

\texttt{Complexity$_{medium}$} & $-4.55$ (1.05)$^*$ & $\phantom{-}14.36$ (5.97)\phantom{$^*$} & $-1.03$ (0.56)\phantom{$^*$} \\ 
\texttt{Complexity$_{large}$} & $-4.54$ (1.04)$^*$ & --- & $-0.51$ (0.72)\phantom{$^*$} \\ 
\texttt{Complexity$_{unknown}$} & $-1.61$ (0.91)\phantom{$^*$} & $-3.23$ (1.87)\phantom{$^*$} & $\phantom{-}0.44$ (0.31)\phantom{$^*$} \\

\texttt{Year} × \texttt{Sample\_Size$_{medium}$} & $-0.59$ (0.31)\phantom{$^*$} & $-0.47$ (0.14)\phantom{$^*$} & $-0.09$ (0.07)\phantom{$^*$} \\ 
\texttt{Year} × \texttt{Sample\_Size$_{large}$} & $\phantom{-}0.05$ (0.26)\phantom{$^*$} & $-1.30$ (0.89)\phantom{$^*$} & $-0.01$ (0.12)\phantom{$^*$} \\ 
\texttt{Year} × \texttt{Sample\_Size$_{unknown}$} & $-0.10$ (0.15)\phantom{$^*$} & $-0.17$ (0.04)\phantom{$^*$} & $\phantom{-}0.04$ (0.07)\phantom{$^*$} \\ 

\texttt{Year} × \texttt{Predictors$_{medium}$} & $-0.04$ (0.08)\phantom{$^*$} & $\phantom{-}1.55$ (0.49)\phantom{$^*$} & $-0.05$ (0.04)\phantom{$^*$} \\ 
\texttt{Year} × \texttt{Predictors$_{large}$} & $-0.15$ (0.18)\phantom{$^*$} & $\phantom{-}0.85$ (0.46)\phantom{$^*$} & $-0.16$ (0.08)\phantom{$^*$} \\ 
\texttt{Year} × \texttt{Predictors$_{unknown}$} & $-0.15$ (0.28)\phantom{$^*$} & $-0.32$ (0.16)\phantom{$^*$} & $-0.03$ (0.11)\phantom{$^*$} \\ 

\texttt{Year} × \texttt{Complexity$_{medium}$} & $\phantom{-}0.14$ (0.16)\phantom{$^*$} & $-1.29$ (0.42)\phantom{$^*$} & $\phantom{-}0.08$ (0.06)\phantom{$^*$} \\ 
\texttt{Year} × \texttt{Complexity$_{large}$} & $\phantom{-}0.38$ (0.09)$^*$ & --- & $\phantom{-}0.04$ (0.07)\phantom{$^*$} \\ 
\texttt{Year} × \texttt{Complexity$_{unknown}$} & $-0.03$ (0.10)\phantom{$^*$} & --- & $-0.06$ (0.03)\phantom{$^*$} \\ 
   \bottomrule
\end{tabular}}
 {\footnotesize\begin{tablenotes}
	\item[]{($^{*}$) indicates statistical significance at the 5\% level.}
	\end{tablenotes}}
\end{threeparttable}
\end{table}

\begin{table}[p]
\centering
\caption{Estimates from mixed-effects models for the EMCS subset analysis.} 
\label{tab:eng_models}
\begin{threeparttable}
{\small
\begin{tabular}{lccc}
   \toprule
   & \textbf{RMSE} & \textbf{AUC} & $\boldsymbol{R}^{\boldsymbol{2}}$\\ 
  \textbf{Term} & \textbf{Estimate (SE)} &  \textbf{Estimate (SE)} & \textbf{Estimate (SE)} \\ 
  \midrule
(Intercept) & $-0.47$ (0.67)\phantom{$^*$} & $\phantom{-}0.26$ (0.25)\phantom{$^*$} & $\phantom{-}0.49$ (2.67)\phantom{$^*$} \\ 

\texttt{Year} & $\phantom{-}0.00$ (0.05)\phantom{$^*$} & $-0.04$ (0.03)\phantom{$^*$} & $-0.01$ (0.11)\phantom{$^*$} \\ 
 
\texttt{Sample\_Size$_{medium}$} & $\phantom{-}1.21$ (0.70)\phantom{$^*$} & --- & $-0.32$ (2.14)\phantom{$^*$} \\
\texttt{Sample\_Size$_{large}$} & $\phantom{-}0.45$ (0.59)\phantom{$^*$} & --- & $-1.67$ (1.31)\phantom{$^*$} \\
\texttt{Sample\_Size$_{unknown}$} & $\phantom{-}0.97$ (1.58)\phantom{$^*$} & --- & $-0.47$ (2.09)\phantom{$^*$} \\

\texttt{Predictors$_{medium}$} & $\phantom{-}0.26$ (0.58)\phantom{$^*$} & --- & $-0.84$ (1.61)\phantom{$^*$} \\ 
\texttt{Predictors$_{large}$} & $\phantom{-}0.60$ (0.92)\phantom{$^*$} & --- & $-3.01$ (1.95)\phantom{$^*$} \\ 
\texttt{Predictors$_{unknown}$} & $-0.94$ (0.59)\phantom{$^*$} & --- & $-0.20$ (1.69)\phantom{$^*$} \\

\texttt{Total\_Params$_{medium}$} & $-0.93$ (3.27)\phantom{$^*$} & --- & --- \\ 
\texttt{Total\_Params$_{large}$} & $-0.15$ (1.83)\phantom{$^*$} & --- & $\phantom{-}4.19$ (1.92)\phantom{$^*$} \\ 
\texttt{Total\_Params$_{unknown}$} & $\phantom{-}0.24$ (1.15)\phantom{$^*$} & --- & $\phantom{-}0.33$ (1.50)\phantom{$^*$} \\

\texttt{Year} × \texttt{Sample\_Size$_{medium}$} & $-0.10$ (0.07)\phantom{$^*$} & --- & --- \\ 
\texttt{Year} × \texttt{Sample\_Size$_{large}$} & $-0.02$ (0.06)\phantom{$^*$} & --- & --- \\ 
\texttt{Year} × \texttt{Sample\_Size$_{unknown}$} & $-0.08$ (0.14)\phantom{$^*$} & --- & --- \\ 

\texttt{Year} × \texttt{Predictors$_{medium}$} & $-0.03$ (0.06)\phantom{$^*$} & --- & --- \\ 
\texttt{Year} × \texttt{Predictors$_{large}$} & $-0.05$ (0.09)\phantom{$^*$} & --- & --- \\ 
\texttt{Year} × \texttt{Predictors$_{unknown}$} & $\phantom{-}0.09$ (0.06)\phantom{$^*$} & --- & --- \\ 

\texttt{Year} × \texttt{Total\_Params$_{medium}$} & $\phantom{-}0.12$ (0.28)\phantom{$^*$} & --- & --- \\ 
\texttt{Year} × \texttt{Total\_Params$_{large}$} & $\phantom{-}0.04$ (0.13)\phantom{$^*$} & --- &  ---\\
   \bottomrule
\end{tabular}}
 {\footnotesize\begin{tablenotes}
	\item[]{($^{*}$) indicates statistical significance at the 5\% level.}
	\end{tablenotes}}
\end{threeparttable}
\end{table}

\begin{table}[p]
\centering
\caption{Estimates from mixed-effects models for the HBS subset analysis.} 
\label{tab:health_models}
\begin{threeparttable}
{\small
\begin{tabular}{lccc}
   \toprule
   & \textbf{RMSE} & \textbf{AUC} & $\boldsymbol{R}^{\boldsymbol{2}}$\\ 
  \textbf{Term} & \textbf{Estimate (SE)} &  \textbf{Estimate (SE)} & \textbf{Estimate (SE)} \\ 
  \midrule
(Intercept) & $-0.06$ (0.02)\phantom{$^*$} & $\phantom{-}0.13$ (0.29)\phantom{$^*$} & --- \\ 

\texttt{Year} & $\phantom{-}0.00$ (0.00)\phantom{$^*$} & $\phantom{-}0.02$ (0.03)\phantom{$^*$} &  ---\\ 

   \bottomrule
\end{tabular}}
 {\footnotesize\begin{tablenotes}
	\item[]{($^{*}$) indicates statistical significance at the 5\% level.}
	\end{tablenotes}}
\end{threeparttable}
\end{table}

\begin{table}[p]
\centering
\caption{Estimates from mixed-effects models for the OTH subset analysis.} 
\label{tab:other_models}
\begin{threeparttable}
{\small
\begin{tabular}{lccc}
   \toprule
   & \textbf{RMSE} & \textbf{AUC} & $\boldsymbol{R}^{\boldsymbol{2}}$\\ 
  \textbf{Term} & \textbf{Estimate (SE)} &  \textbf{Estimate (SE)} & \textbf{Estimate (SE)} \\ 
  \midrule
(Intercept) & $-2.43$ (4.50)\phantom{$^*$} & $-0.18$ (0.73)\phantom{$^*$} & $\phantom{-}1.25$ (2.37) \\ 

\texttt{Year} & $\phantom{-}0.25$ (0.43)\phantom{$^*$} & $\phantom{-}0.01$ (0.04)\phantom{$^*$} & $-0.01$ (0.28) \\ 
 
\texttt{Sample\_Size$_{medium}$} & $\phantom{-}0.38$ (0.44)\phantom{$^*$} & $-1.56$ (2.99)\phantom{$^*$} & $\phantom{-}0.64$ (2.56) \\
\texttt{Sample\_Size$_{large}$} & $\phantom{-}0.06$ (0.45)\phantom{$^*$} & $-0.30$ (0.65)\phantom{$^*$} & $-0.54$ (2.32) \\
\texttt{Sample\_Size$_{unknown}$} & $\phantom{-}0.04$ (0.28)\phantom{$^*$} & --- & $\phantom{-}0.84$ (1.08) \\

\texttt{Predictors$_{medium}$} & $-0.12$ (0.34)\phantom{$^*$} & --- & --- \\ 
\texttt{Predictors$_{large}$} & $-0.15$ (0.32)\phantom{$^*$} & --- & $\phantom{-}$0.31 (0.93) \\ 
\texttt{Predictors$_{unknown}$} & $-0.03$ (0.23)\phantom{$^*$} & --- & $\phantom{-}$0.32 (0.98) \\

\texttt{Total\_Params$_{medium}$} & $\phantom{-}2.72$ (4.37)\phantom{$^*$} & --- & $-0.90$ (1.70) \\ 
\texttt{Total\_Params$_{large}$} & $-0.02$ (5.11)\phantom{$^*$} & --- & $-0.59$ (2.41) \\ 
\texttt{Total\_Params$_{unknown}$} & $\phantom{-}2.12$ (4.55)\phantom{$^*$} & --- & $-1.78$ (1.11) \\

\texttt{Year} × \texttt{Total\_Params$_{medium}$} & $-0.27$ (0.43)\phantom{$^*$} & --- & --- \\ 
\texttt{Year} × \texttt{Total\_Params$_{large}$} & $-0.01$ (0.50)\phantom{$^*$} & --- & --- \\ 
\texttt{Year} × \texttt{Total\_Params$_{unknown}$} & $-0.21$ (0.44)\phantom{$^*$} & --- & --- \\ 

\texttt{Year} × \texttt{Sample\_Size$_{medium}$} & --- & $-0.56$ (0.73)\phantom{$^*$} & --- \\ 

   \bottomrule
\end{tabular}}
 {\footnotesize\begin{tablenotes}
	\item[]{($^{*}$) indicates statistical significance at the 5\% level.}
	\end{tablenotes}}
\end{threeparttable}
\end{table}

\begin{table}[p]
\centering
\caption{Estimates from mixed-effects models weighted by sample size.}
\label{tab:sens_weighted}
\begin{threeparttable}
{\small
\begin{tabular}{lccc}
\toprule
 & \textbf{RMSE} & \textbf{AUC} & $\mathbf{R}^2$ \\
\textbf{Term} & \textbf{Estimate (SE)} & \textbf{Estimate (SE)} & \textbf{Estimate (SE)} \\
\midrule
(Intercept) 
 & $\phantom{-}$0.14 (0.26) & $-0.02$ (0.44)\phantom{$^*$} & $-0.90$ (0.95) \\

\texttt{Year} 
 & $-0.03$ (0.03) & $\phantom{-}$0.07 (0.07)\phantom{$^*$} & $\phantom{-}$0.07 (0.08) \\

\texttt{Area$_{\text{Env/Agri}}$}
 & $-0.43$ (0.44) & $\phantom{-}$0.07 (0.50)\phantom{$^*$} & $\phantom{-}$0.64 (0.81) \\

\texttt{Area$_{\text{Health/Bio}}$}
 & $-0.30$ (0.62) & $\phantom{-}$0.00 (0.51)\phantom{$^*$} & $-0.18$ (0.79) \\

\texttt{Area$_{\text{Other}}$}
 & $-0.30$ (0.55) & $-0.11$ (0.53)\phantom{$^*$} & $\phantom{-}$0.33 (1.06) \\

\texttt{Predictors$_{\text{medium}}$}
 & $\phantom{-}$0.02 (0.44) & $-0.28$ (0.22)\phantom{$^*$} & $\phantom{-}$0.81 (0.56) \\

\texttt{Predictors$_{\text{large}}$}
 & $\phantom{-}$0.31 (0.45) & $\phantom{-}$0.55 (0.42)\phantom{$^*$} & $\phantom{-}$0.78 (1.17) \\

\texttt{Year} × \texttt{Area$_{\text{Env/Agri}}$}
 & $\phantom{-}$0.04 (0.04) & $-0.05$ (0.06)\phantom{$^*$} & $-0.06$ (0.07) \\

\texttt{Year} × \texttt{Area$_{\text{Health/Bio}}$}
 & $\phantom{-}$0.03 (0.06) & $\phantom{-}$0.06 (0.06)\phantom{$^*$} & --- \\

\texttt{Year} × \texttt{Area$_{\text{Other}}$}
 & $\phantom{-}$0.03 (0.06) & $\phantom{-}$0.00 (0.06)\phantom{$^*$} & $-0.01$ (0.10) \\

\texttt{Year} × \texttt{Predictors$_{\text{medium}}$}
 & $\phantom{-}$0.02 (0.04) & $-0.05$ (0.05)\phantom{$^*$} & $-0.04$ (0.06) \\

\texttt{Year} × \texttt{Predictors$_{\text{large}}$}
 & $-0.01$ (0.04) & $-0.13$ (0.06)$^{*}$ & $-0.06$ (0.11) \\

\midrule
$\sigma^2_{\text{paper}}$ 
 & $\phantom{-}$0.10 \phantom{(0.00)} & $\phantom{-}$0.14 \phantom{$(0.00)^*$} & $\phantom{-}$0.43 \phantom{(0.00)} \\
$\sigma^2_{\text{residual}}$ 
 & 229.85 \phantom{(0.00)} & 79.49 \phantom{(0.00)} & $\phantom{-}$2.40 \phantom{(0.00)}\\
ICC 
 & $\phantom{-}$0.00 \phantom{(0.00)} & $\phantom{-}$0.97 \phantom{$(0.00)^*$} & $\phantom{-}$0.15 \phantom{(0.00)} \\
\bottomrule
\end{tabular}}

\begin{tablenotes}
\footnotesize
\item[] (*) indicates statistical significance at the 5\% level.  
\end{tablenotes}
\end{threeparttable}
\end{table}

\end{document}